\documentclass[10pt,twocolumn,letterpaper]{article}

\usepackage{iccv}
\usepackage{times}
\usepackage{epsfig}
\usepackage{graphicx}
\usepackage{amsmath}
\usepackage{amssymb}

\usepackage{booktabs}
\usepackage{cite}
\usepackage{multirow}
\usepackage{subcaption}

\usepackage[T1]{fontenc}
\usepackage[utf8]{inputenc}
\usepackage{authblk}

\usepackage[accsupp]{axessibility} % Improves PDF readability for those with disabilities.

\usepackage[breaklinks=true,bookmarks=false]{hyperref}

\iccvfinalcopy % *** Uncomment this line for the final submission

\def\iccvPaperID{10450} % *** Enter the ICCV Paper ID here

\ificcvfinal\fi

\begin{document}

%%%%%%%%% TITLE
\title{SYENet: A Simple Yet Effective Network for Multiple Low-Level Vision Tasks with Real-time Performance on Mobile Device}

\author[1,2]{Weiran Gou$^*$}
\author[1,2]{Ziyao Yi$^*$}
\author[1,2]{Yan Xiang$^*$}
\author[1,2]{Shaoqing Li}
\author[1,2]{Zibin Liu}
\author[1,2]{Dehui Kong}
\author[1,2]{Ke Xu$^\dagger$}



% \and
\affil[1]{State Key Laboratory of Mobile Network and Mobile Multimedia Technology}
\affil[2]{Sanechips Technology, Chengdu, China}
\affil[ ]{\tt\small \{gou.weiran, yi.ziyao, xiang.yan1, li.shaoqing1, liu.zibin, kong.dehui, xu.kevin\}@sanechips.com.cn}
\affil[ ]{\href{https://github.com/sanechips-multimedia/syenet}{\textcolor{magenta}{\tt\small{https://github.com/sanechips-multimedia/syenet}}}}

\maketitle
\ificcvfinal\thispagestyle{empty}\fi

\def\thefootnote{$*$}\footnotetext{These authors contributed equally to this work.}\def\thefootnote{\arabic{footnote}}
\def\thefootnote{$\dagger$}\footnotetext{Corresponding author: Ke Xu(xu.kevin@sanechips.com.cn)}\def\thefootnote{\arabic{footnote}}
% \def\thefootnote{$1$}\footnotetext{Codes are available at \href{https://github.com/sanechips-multimedia/syenet}{\textcolor{magenta}{\tt\small{https://github.com/sanechips-multimedia/syenet}}}}\def\thefootnote{\arabic{footnote}}

%text text text\footnote{normal footnote}

%%%%%%%%% ABSTRACT
\begin{abstract}

With the rapid development of AI hardware accelerators, applying deep learning-based algorithms to solve various low-level vision tasks on mobile devices has gradually become possible. However, two main problems still need to be solved: task-specific algorithms make it difficult to integrate them into a single neural network architecture, and large amounts of parameters make it difficult to achieve real-time inference. To tackle these problems, we propose a novel network, SYENet, with only $~$6K parameters, to handle multiple low-level vision tasks on mobile devices in a real-time manner. The SYENet consists of two asymmetrical branches with simple building blocks. To effectively connect the results by asymmetrical branches, a Quadratic Connection Unit(\textbf{QCU}) is proposed. Furthermore, to improve performance, a new Outlier-Aware Loss is proposed to process the image. The proposed method proves its superior performance with the best PSNR as compared with other networks in real-time applications such as Image Signal Processing(ISP), Low-Light Enhancement(LLE), and Super-Resolution(SR) with 2K60FPS throughput on Qualcomm 8 Gen 1 mobile SoC(System-on-Chip). Particularly, for ISP task, SYENet got the highest score in MAI 2022 Learned Smartphone ISP challenge. 

% $^1$ % Codes are available at \href{https://github.com/sanechips-multimedia/syenet}{\textcolor{magenta}{\tt\small{https://github.com/sanechips-multimedia/syenet}}}.

\end{abstract}

%%%%%%%%% BODY TEXT
\section{Introduction} 
\label{sec:introduction}

In recent years, with the thriving development of AI accelerators\cite{8662396,8662476}, such as Neural Processor Units(NPUs) or Graphic Processor Units(GPUs), AI algorithms can be deployed on mobile devices and achieved great success\cite{Zhang_2018_CVPR,Ma_2018_ECCV,9389905,xiong2021mobiledets}. Many mobile SoCs, especially those designed for smartphone, tablet, and in-vehicle infotainment systems, require superior visual quality processing, which cannot be achieved without leveraging deep networks such as ISP\cite{ignatov2020replacing,Ignatov2021maireport}, LLE\cite{chen2018learning}, and SR\cite{chu2022nafssr,chen2021pre,dai2019second}. However, due to the tight hardware constraints such as power and computing resources, deploying these algorithms on mobile devices still has several issues as follows.  

\begin{figure}
  \centering
  
  \begin{subfigure}{0.9\linewidth}
    \includegraphics[width=1.0\linewidth]{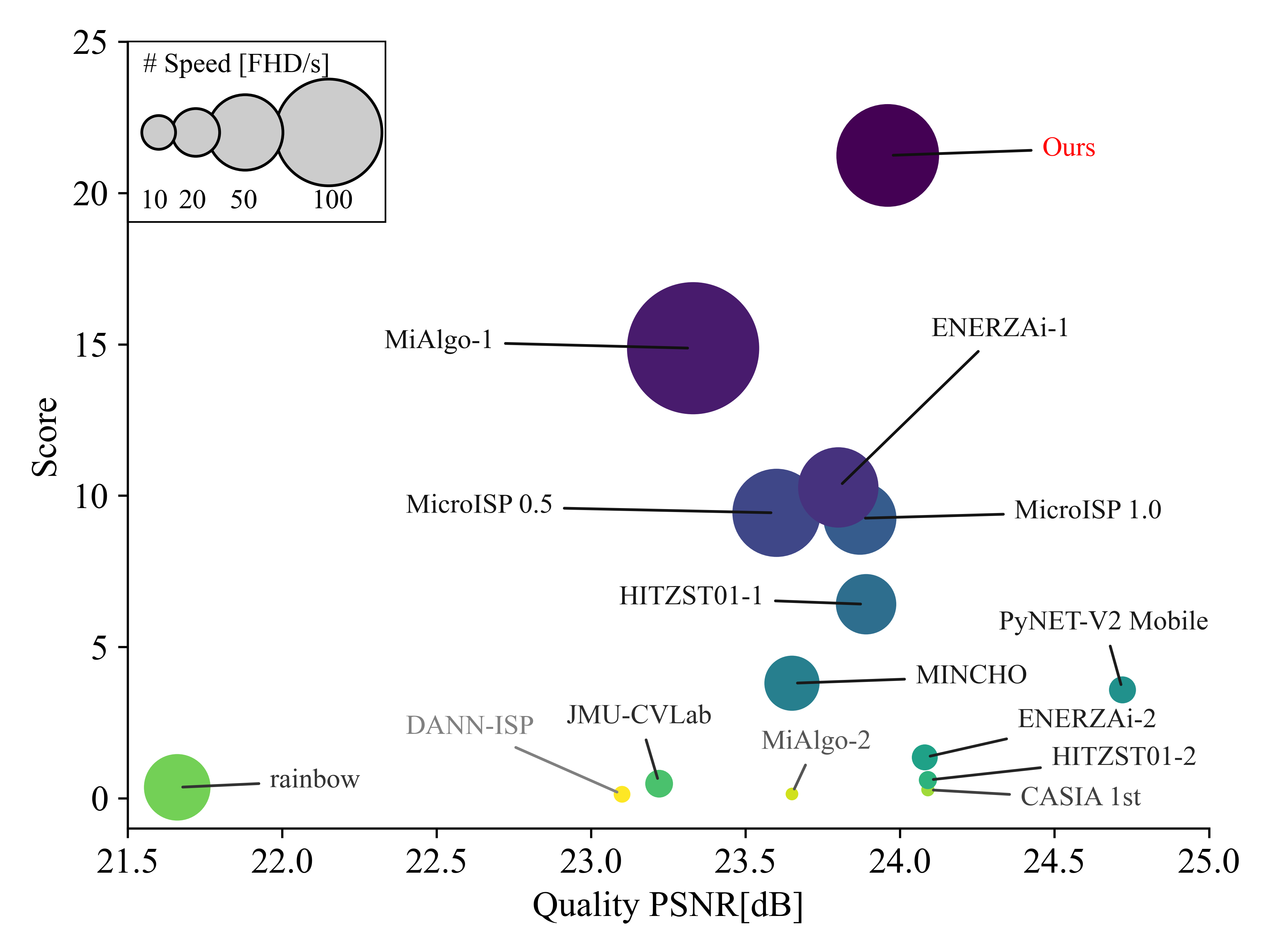}
    \caption{ISP}
    \label{ivq-isp}
  \end{subfigure}
  %\hfill
  
  %\centering
  %\begin{subfigure}{0.9\linewidth}
  %  \centering
  %  \includegraphics[width=1.0\linewidth]{Figures/SRX2Comp.png}
  %  \caption{SR with scale factor $\times$2}
  %  \label{ivq-sr2}
  %\end{subfigure}

  \centering
  \begin{subfigure}{0.9\linewidth}
    \centering
    \includegraphics[width=1.0\linewidth]{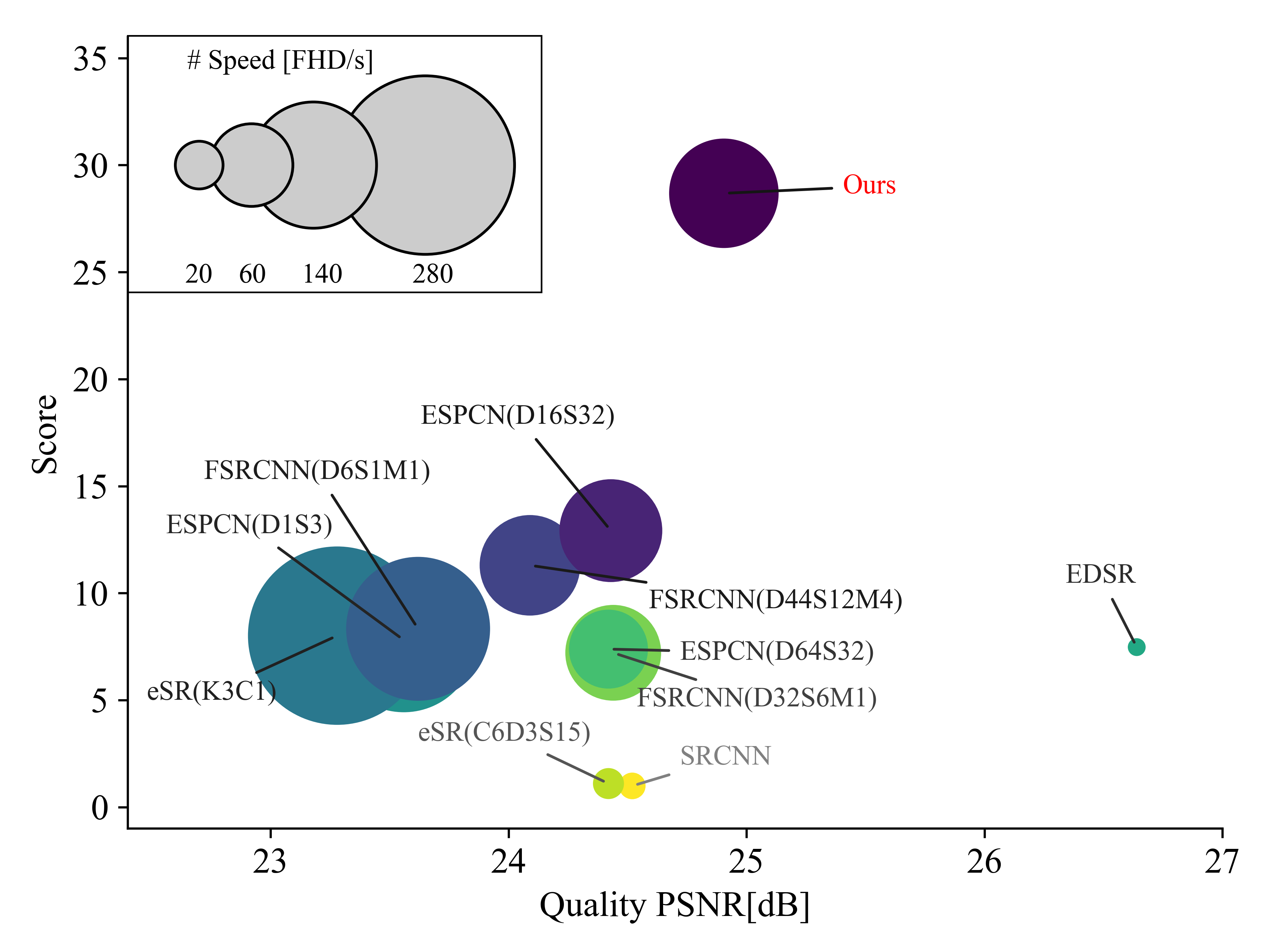}
    \caption{SR with scale factor $\times$4}
    \label{ivq-sr4}
  \end{subfigure}

    %(c)SR$\times4$
  \caption{Comparison about different issues (a)ISP (b)SR$\times4$ upon comprehensive score versus quantitative measurements by SOTA models. The size of the model represents the inference speed. The Score equation is in Eq. \ref{eq-score} by the MAI Challenge\cite{Ignatov2021maireport}. Our method shows superior comprehensive performance upon image quality, inference speed, and the score involving both factors.} 
  \label{ivq}
  %\vspace{-8mm}
\end{figure}

The first issue concerns real-time processing. Usually, these low-level vision tasks require a 2K60FPS or even higher real-time performance to satisfy the viewer's needs. Although the State-of-the-Arts(SOTAs)\cite{chen2021pre,ignatov2020replacing,dai2019second,wang2022low} dealing with similar tasks have boosted the performance, they increased the numbers of parameters and computational cost drastically, which cannot satisfy real-time inference deployment even on powerful hardware such as server-level processors. Moreover, compared with high-level tasks\cite{Zhang_2018_CVPR, xiong2021mobiledets}, where the input images could be resized into a lower resolution such as $128\times128$ or $256\times256$ without noticeable effects, low-level vision tasks cannot do the same thing as their preliminary goal is to improve the human visual quality. A more detailed discussion about the constraints of low-level vision tasks is in Appendix \ref{sec:app-constraints}.

The second issue is related to hardware resources on mobile devices such as Qualcomm's Snapdragon. As compared with server-level Central Processing Unit(CPU) or GPU, mobile SoC usually has limited computing resources such as multiplication-and-accumulation units, limited memory bandwidth, and limited power consumption budget. Unfortunately, most low-level vision algorithms are task-specific\cite{dai2019second,lim2017enhanced,chen2018learning,ignatov2020replacing} and independent to each other, which makes it difficult to merge into a single architecture. To make things worse, many advanced operators, such as deformable convolution\cite{zhu2019deformable} and 3D-convolution\cite{maturana2015voxnet}, cannot be directly applied on mobile devices, which further leads to performance degradation. Therefore, as already proved in high-level vision tasks and NLP\cite{MaskedAutoencoders2021,he2019moco,chen2020mocov2,brown2020language}, building a simple yet unified network architecture is the best choice for low-level vision tasks running on limited computing resources. Although there are excellent multiple low-level vision works like\cite{chen2021pre,liang2021swinir,chen2022simple}, they are not feasible for deployment on mobile devices due to their hardware complexity.

Several lightweight models \cite{lee2019mobisr,liu2021splitsr,ayazoglu2021extremely,wang2020practical} were already proposed with a relatively small number of parameters to achieve a reasonable performance. Unfortunately, their implementations cannot satisfy real-time requirements such as 2K60FPS. To the best of our knowledge, there is still no prior work for the multiple low-level vision tasks in a single network architecture.  

In this paper, we propose a new architecture SYENet, which can solve multiple low-level vision tasks with 2K60FPS on a mobile device such as Qualcomm's 8 Gen 1. We first decompose the low-level vision into two sub-tasks, which are texture generation and pattern classification. We then leverage two asymmetric branches to handle each task and a Quadratic Connection Unit(\textbf{QCU}) to connect the outputs to enlarge the representational power. Furthermore, the network replaces ordinary convolution with revised re-parameterized convolution to boost the performance without increasing inference time, and Channel Attention(CA) is utilized for enhancement by global information. In addition, we propose Outlier-Aware Loss by involving global information and
putting more focus on the outliers of the prediction for improving the performance. The proposed network achieves SOTA performance, as compared with other methods on low-level tasks. The comprehensive performance evaluation of SR, LLE and ISP tasks are shown in Table \ref{table-sr}, \ref{table-lle}, and \ref{table-isp}, respectively.

The contributions of this paper can be summarized in three aspects:
\begin{enumerate}

\iffalse
\item We theoretically analyze low-level vision tasks and propose a simple network architecture based on asymmetric branch, \textbf{QCU}, revised re-parameter convolution, and channel attention to solving multiple low-level vision tasks.

\item A new loss function termed \textbf{Outlier-Aware Loss} is proposed to calculate the loss to improve the performance. In addition, the \textbf{QCU} is proposed to fuse the outputs from asymmetric branches. Both \textbf{Outlier-Aware Loss} and \textbf{QCU} improve the performance of the network.
\fi

\item We propose that asymmetric branches fused with Quadratic Connections Unit(\textbf{QCU}) is an effective method for solving multiple low-level vision tasks due to its ability to enlarge the representation power with modicum parameter count. Building upon this structure, we introduce SYENet, which incorporates revised reparameterized convolutions and channel attention to enhance performance without sacrificing speed.

\item A new loss function termed \textbf{Outlier-Aware Loss} is proposed for better training by leveraging global information and prioritizing outliers, the poorly predicted pixels. 

% We also demonstrate the generalizability of this approach by achieving improved performance on other models.

\item Compared with other studies, our network has a superior performance according to the evaluation metrics in MAI Challenge \cite{mai2021isp}, which reflects both the image quality and efficiency as shown in Fig. \ref{ivq}. 

\end{enumerate}

%------------------------------------------------------------------------

\section{Related work}
\label{sec:related-work}

\subsection{Low-level vision}

Low-level vision techniques are generally required in a variety of applications to improve image and video quality. It could be defined as finding the best mapping between input and output images. In this section, we mainly discuss three widely used low-level vision tasks, which are super-resolution \textbf{SR}, end-to-end image signal processing \textbf{ISP}, and low-light enhancement \textbf{LLE}. 

\textbf{Super resolution}: Convolution Neural Network(CNN) are widely used in SR algorithms. From the very first model SRCNN\cite{dong2014learning} to EDSR\cite{lim2017enhanced}, ESPCN\cite{shi2016real}, FEQE\cite{vu2018fast} and VDSR\cite{kim2016accurate} .etc, CNNs significantly improve\cite{zhang2018residual, dai2019second, mei2021image} SR performance and try to reduce the computational complexity. Special building blocks such as residual block\cite{he2016deep,zhang2018image,dai2019second} and deformable convolution\cite{wang2019edvr, kong2020efficient} are also used to improve visual quality. Transformer-based SR models such as SwinSR\cite{liang2021swinir} and IPT\cite{chen2021pre} show significant improvements compared to traditional CNN-based models.

\textbf{End-to-end ISP}: HighEr-Resolution Network(HERN)\cite{mei2019higer} employs a two-branch structure to combine features of different scales to help conduct the tasks of demosaicing and image enhancement. PyNet\cite{ignatov2020replacing} achieves similar performance as compared with the most sophisticated traditional ISP pipelines. AWNet\cite{dai2020awnet} introduces attention mechanism, and wavelet transform for learning-based ISP network, which significantly improves image quality due to a large receptive field. Focusing on the color inconsistency issue that exists between input raw and output sRGB images, Zhang\cite{zhang2021raw2srgb} designs a joint learning network. Similarly, from the perspective of solving noise discrepancy, Cao\cite{cao2021pseudo} introduces a pseudo-ISP, utilizing unpaired learning algorithm.

\textbf{Low-light enhancement}: Some end-to-end RAW-to-RGB LLE methods \cite{hasinoff@burst,Alpher17,Alpher18} employ the color shuffle operator in the front of the network. In the sRGB domain, with the advantage of being interpretable, many researchers focus on the decomposition method for LLE task, enhancing neural network designs and additional regularization as used in de-haze and de-noise \cite{Alpher19,Alpher20,Alpher21,Alpher22,Alpher23}. Based on the non-local evaluation, normal light output can be obtained through a global awareness or generation method \cite{Alpher24,Alpher25,Alpher26}.

\begin{figure*}
  \centering
  
  \begin{subfigure}{1.0\linewidth}
    \includegraphics[width=1.0\linewidth]{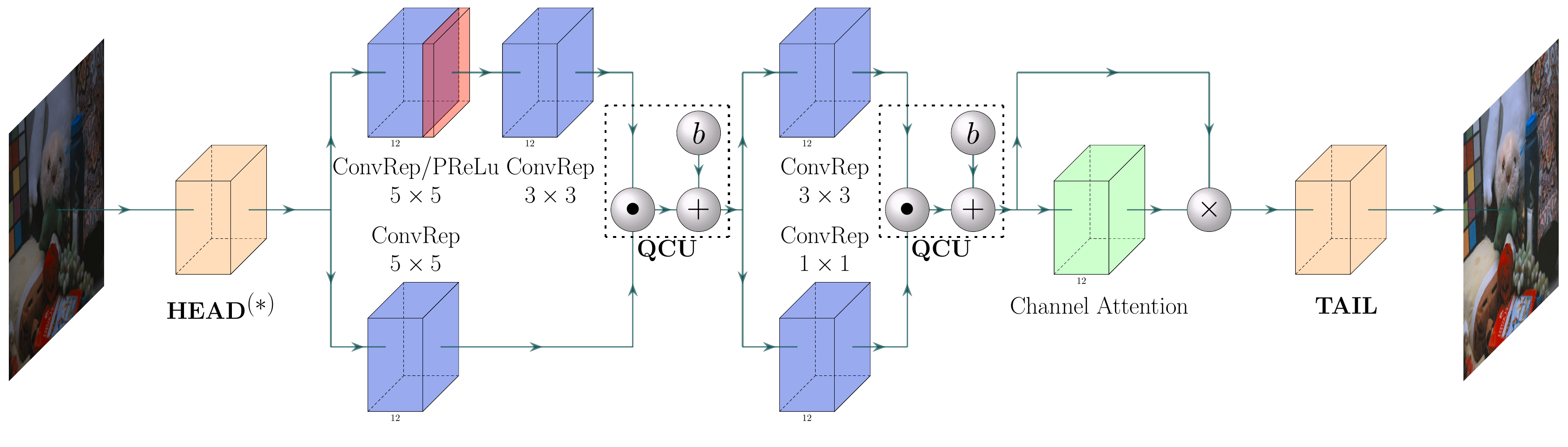}
    \caption{Overall Architecture of SYENet: two $\odot$ operations are element-wise multiplications and $\otimes$ operation is channel-wise multiplication (* means some tasks may not require a head block to process). After the reparameterization, SYENet consists of 6 convolutions with only 5K parameters, excluding head and tail blocks.}
    \label{sye-archi}
  \end{subfigure}
  %\vspace{-10mm}
  
  \hfill
  
  \centering
  \begin{subfigure}{1.0\linewidth}
    \centering
    \includegraphics[width=1.0\linewidth]{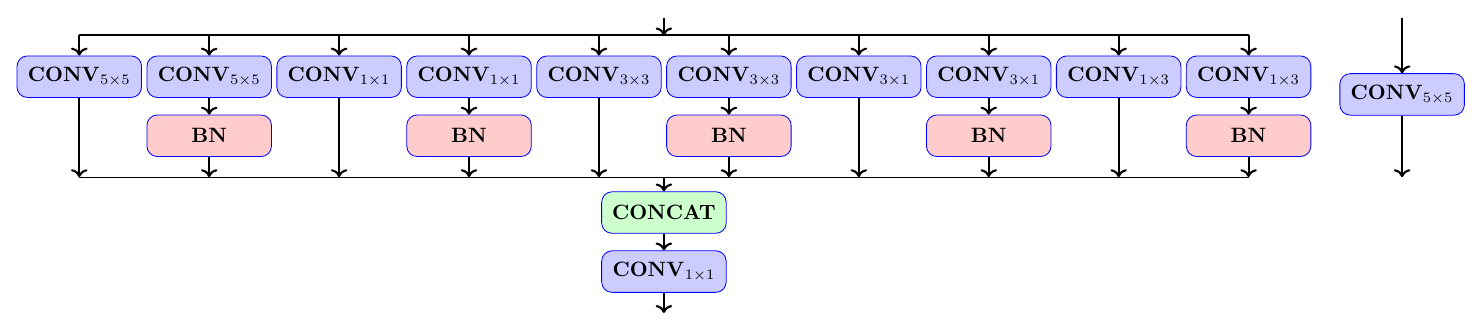}
    \caption{ConvRep block during training(left) and inference(right) phase, the training branches can be specifically designed for different requirements and applications.}
    \label{sye-rep}
  \end{subfigure}
  %\vspace{-7mm}
  \caption{Architecture of SYENet and the structure of ConvRep block in training (left) as well as inference (right) phase}
  \label{sye}
%\vspace{-7mm}
\end{figure*}

%%%%%%%%%%%%%%%%%%%%%%%%%%%%%%%%%%%%%%%%%%%%%%%%%%%%%%%%%%%%%%%%%%%%%%%%

%-------------------------------------------------------------------------

\subsection{Mobile devices implementation}

The SOTA networks for solving low-level vision problems show increasingly good performance. However, most of them are too computationally expensive, and hence it is tough to implement those algorithms in mobile devices without a powerful GPU. Meanwhile, some research about compact and effective network were carried out. Wang\cite{wang2020practical} proposed a lightweight U-shape network to support denoising operations on mobile platforms. MobiSR\cite{lee2019mobisr} with model compression methods applies two networks focusing on latency as well as quality to guarantee efficiency. SplitSR\cite{liu2021splitsr} reached 5 times faster inference using lightweight residual block, and XLSR\cite{ayazoglu2021extremely} applies deep roots module\cite{ioannou2017deeproots} into SISR issue demonstrating the same performance of VDSR\cite{kim2016accurate} using 30 times fewer parameters. Unfortunately, however, lightweight networks still preserve millions of parameters, which is far from the real-time application of 2K60FPS in mobile devices. 

\subsection{Re-parameterization}
 
% \textbf{Re-parameterization}: 

Re-parameterization is the approach for structural simplification using re-parameterized blocks, which is complicated during training but simplified during inference with the equivalent forward results. ACNet\cite{ding2019acnet} inspired by the idea of convolution factorization, introduces asymmetric convolution block(ACB), which slightly improves performance and significantly reduces the computational cost. RepVGG\cite{ding2021repvgg} which is inspired by ResNet\cite{he2016resnet} applies RepVGG block with skip connections to replace the normal single convolution block. Later on, RepOptVGG\cite{ding2022reopt} proposed to use the re-parameterized optimizer to replace the re-parameterized network architecture, which could even additionally dismiss the complexity in the training phase compared with RepVGG. In this study, the technique of re-parameterization shall be utilized to help SYENet to accelerate the inference.

%-------------------------------------------------------------------------
\section{Method}
\label{sec:SYENet}

As the target platform for SYENet is mobile device, which has very limited hardware resources compared to cloud computing, each building block of SYENet should be carefully designed to reduce computation complexity while retaining the desired performance.

\subsection{Texture generation and pattern selection}

To reconstruct the desired images from the degraded input, texture and pixel pattern, which are compact representations and useful features, should be extracted and processed. The texture feature is the base for pixel prediction in SYENet. Pattern information reflected by color provides each pixel with classification information and is utilized to guide pixel prediction. Apparently, extracting the texture features as the regression task requires a deeper network for a larger receptive field than that of pattern information extraction as the classification task. Therefore, we use the asymmetric module with two branches for these tasks. The texture generation branch is designed to have two layers of convolutions, while the pattern selection branch only has one. For the same reason, the second asymmetric block is designed to have two branches with a $3\times 3$ and $1\times 1$ kernel convolution, respectively. The output of the two branches is shown in Fig. \ref{fmap-show}, and more examples can be found in Appendix \ref{apdx:fmap}.

\subsection{Quadratic Connection Unit (\textbf{QCU}): improving the capability of fitting arbitrary models}

\begin{figure}[t]
  \centering
    \includegraphics[width=0.95\linewidth]{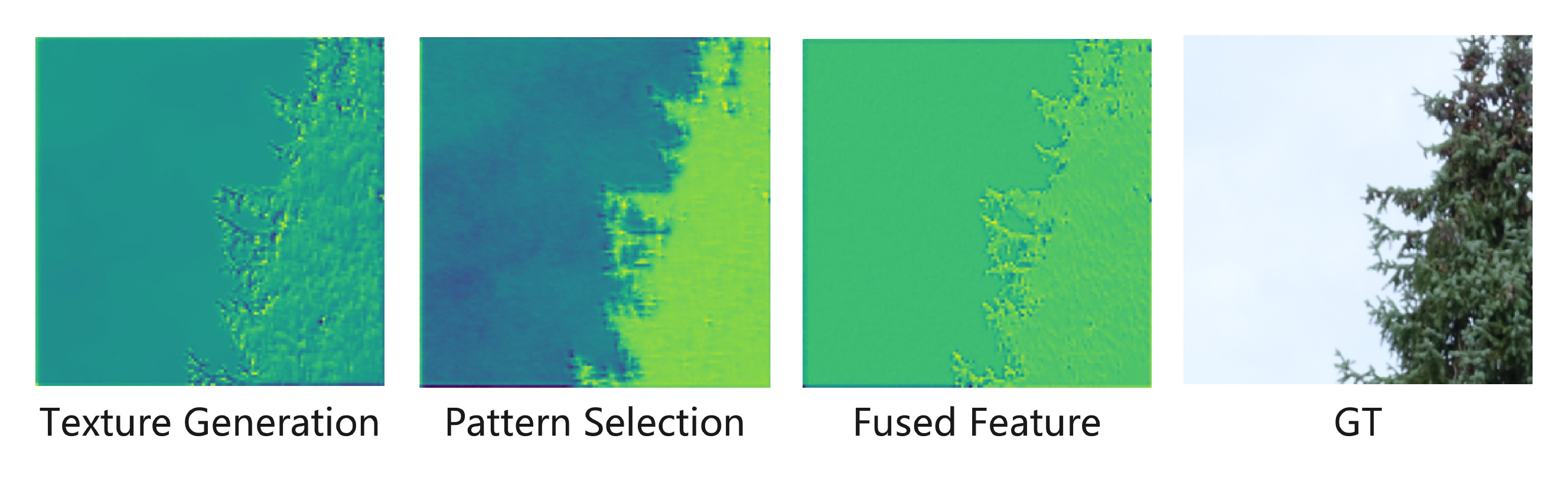}
    %\vspace{-4mm}
    \caption{Complex texture feature with many details, simple pattern classification focusing on labeling and clustering pixels, fused results, and the ground truth}
    \label{fmap-show}
    %\vspace{-5mm}
\end{figure}

\begin{figure}[t]
  \centering
    \includegraphics[width=1.0\linewidth]{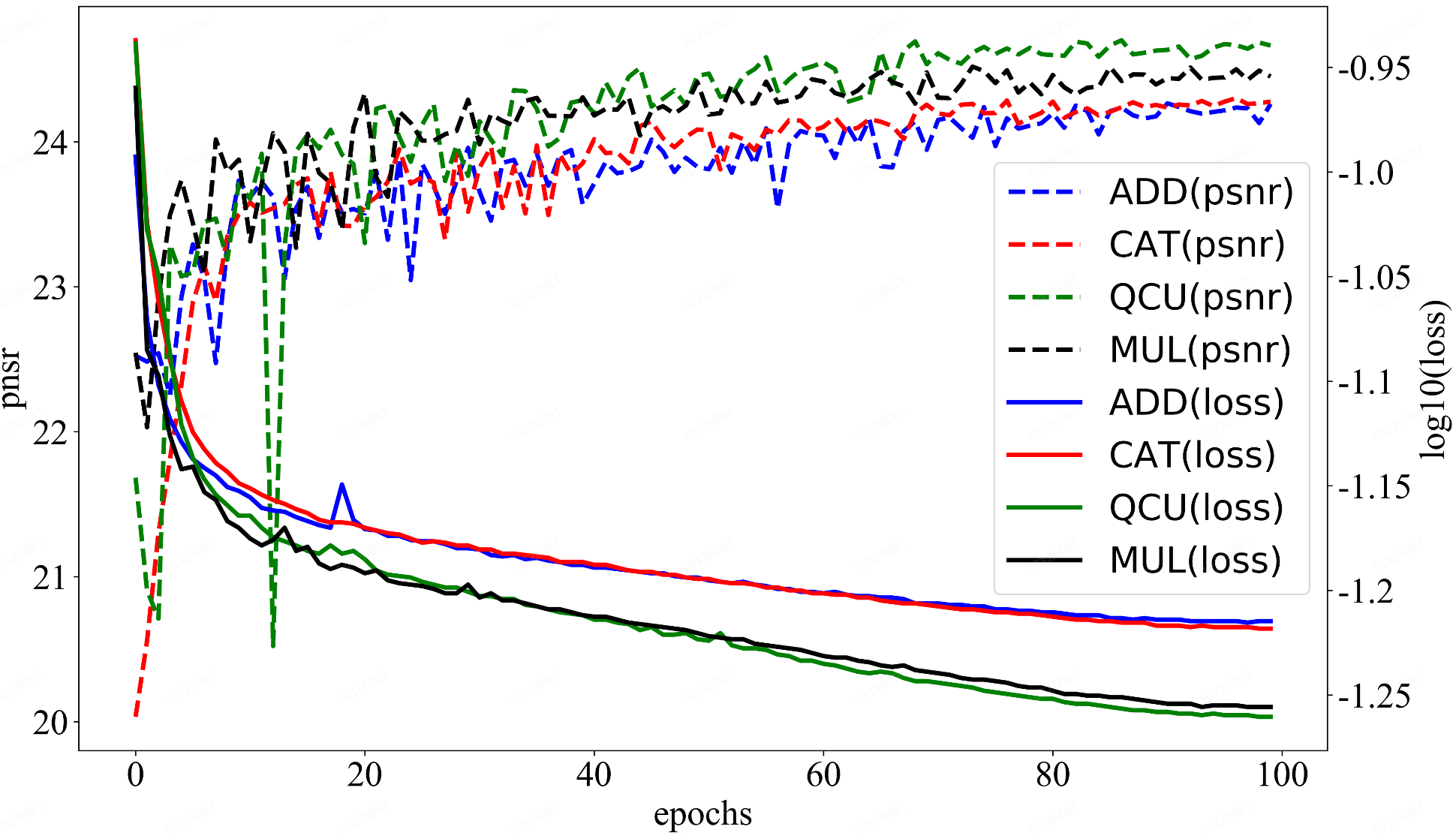}
    %\vspace{-5mm}
    \caption{Faster convergence and higher PSNR by QCU compared with various fusion methods (addition(ADD), concatenation(CAT), multiplication(MUL)) in training SYENet for LLE task in LoL dataset. The QCU reaches higher PSNR and lower loss during the training.}
    \label{fusion-comp}
    %\vspace{-5mm}
\end{figure}

Typically, in the previous multi-branch networks, the fusion of outputs by different branches could be done by concatenation \cite{szegedy@going,ayazoglu2021extremely} or element-wise addition followed by activation function \cite{gharbi2017hdrnet,ding2019acnet}. In this study, in order to effectively improve the representational power, a Quadratic Connection Unit (\textbf{QCU}), as Eq. \ref{eq-qcu} where $\odot$ is an element-wise multiplication and $\oplus$ is element-wise addition, is employed for the fusion of the results by two branches $F_1$ and $F_2$. In big models with numerous channels, employing \textbf{QCU} may not make a difference because big models already have powerful expressiveness. However, for small models like SYENet this revision is rather vital. 

\begin{equation}
  \mathbf{QCU}(F_1,F_2)= (F_1 \odot F_2) \oplus \mathcal{B}
  \label{eq-qcu}
\end{equation}

The formulation of $F_1$ and $F_2$ after re-parameterization shall be represented as linear form $KX+B$ due to convolution being linear transformation, so that the multiplied output should be in the quadratic form as $(\hat{K}X+\hat{B})(\tilde{K}X+\tilde{B})$. In addition, NAFNet\cite{chen2022simple} revealed that activation could be replaced by multiplication in terms of providing nonlinearity towards the network. 

However, we find that there exists the constraint or drawback of the above quadratic form by pure multiplication that the function must pass through two fixed position sets $(-\hat{B}/\hat{K},0)$ and $(-\tilde{B}/\tilde{K},0)$. Meanwhile, multiplication rather than addition could more easily enhance the influence of perturbations, which impairs robustness. To fix the two issues mentioned above, we add an element-wise learnable bias $\mathcal{B}$ to the fused output, which can impressively convert the expression to a more general form as $K_2X^2+K_1X+B$.

\subsection{Outlier-Aware Loss: putting more focus on erroneously predicted pixels}

% $\mathcal{L}_{1}$ loss is widely exploited in image restoration tasks. However, since $\mathcal{L}_{1}$ loss is the pixel-to-pixel distance in the absence of global information, it dismisses the fact that viewers may focus more on the overall perceptual quality instead of individual pixels. Hence 

\begin{figure*}[t]
  \centering
    \includegraphics[width=1.0\linewidth]{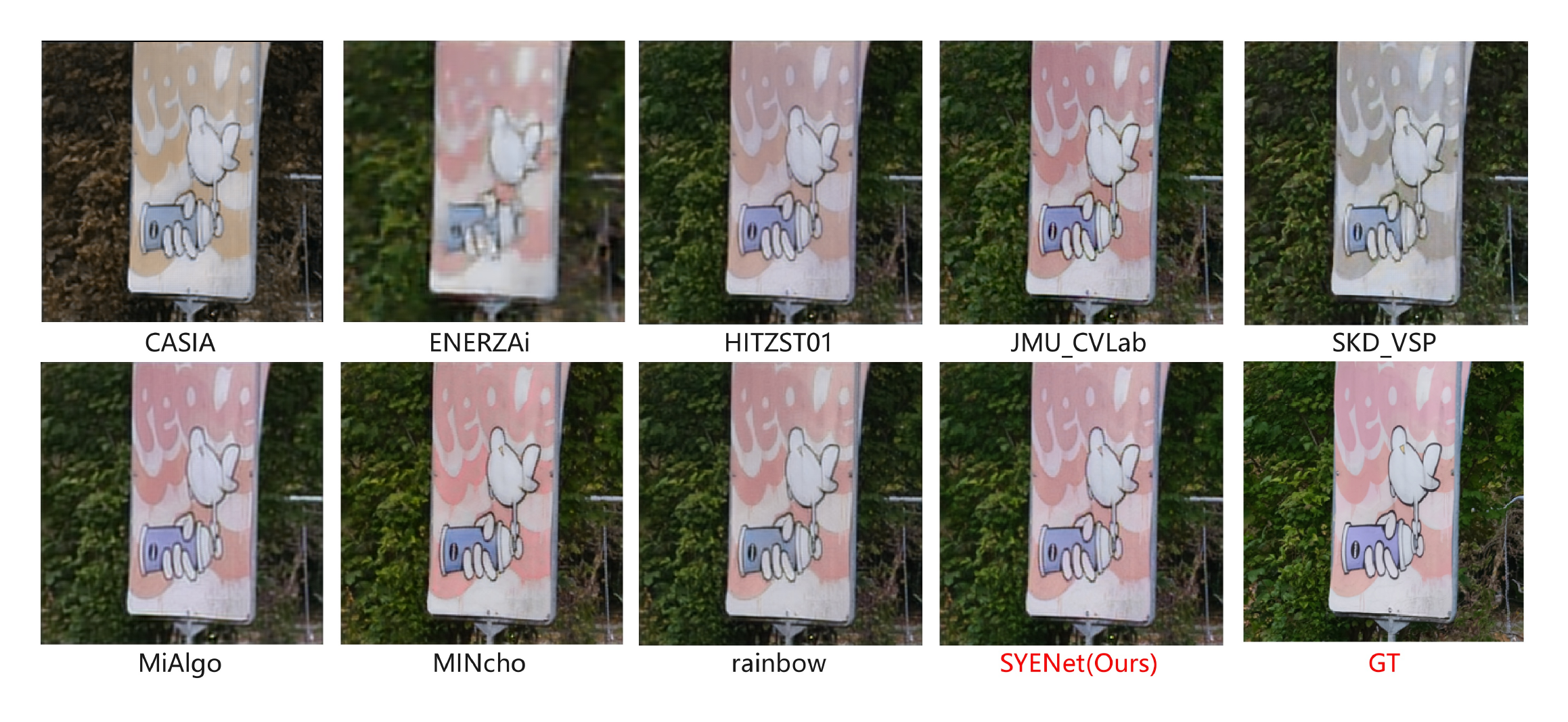}
    %\vspace{-11mm}
   \caption{Image signal processing comparisons with models from participators of MAI 2022 Challenge: Our model shows competitive performance compared with other efficient small networks, and the detailed quantitative comparisons are in Table \ref{table-isp}. Even though the PSNR of our method is not the highest, the comprehensive performance measured by score (Eq. \ref{eq-score}) is the highest. More comparisons of qualitative results are presented in the Appendix \ref{apdx:isp-comp}.}
   \label{isp-comp}
%\vspace{-5mm}
\end{figure*}

In this study, applying the idea of Focal Loss\cite{lin2017focalloss} to regression problem, we propose a new loss function termed Outlier-Aware Loss $\mathcal{L}_{OA}$, as shown in Eq. \ref{eq-lp}, involving global information and putting more focus on the pixels that are badly predicted as the outliers. In Eq. \ref{eq-delta}, $\Delta$ is the difference between ground truth $I^{GT}$ and the output by SYENet $I^{SYE}$ in matrix form, and $\delta_{i,j}$ is the value of $\Delta$ in position $(i,j)$. In Eq. \ref{eq-lp}, $H$ and $W$ are the output height and width. $\mu$ and $\sigma^2$, as the global information, are the mean and variance of $\Delta$. $b$ is the scale parameter defined by $2b^2=\sigma^2$. $\alpha$ is a tunable hyperparameter assigned by the user. Compared with $\mathcal{L}_1$ loss, the loss in pixel $(i,j)$ is multiplied by a weight $W_{i,j} = 1-e^{-\alpha{|\delta_{i,j}-\mu|}^p/b}$. $W_{i,j}$ is proportional to $|\delta_{i,j}-\mu|$ and allows the model to focus on hard, erroneously predicted pixels. $p$ is the norm number and is normally set to be 1 in low-level vision tasks implying the original loss to be optimized by $W$ is $\mathcal{L}_1$ loss. Moreover, as shown in Table \ref{table-abs}, Fig. \ref{lle-comp}, Fig. \ref{sr-comp}, and Fig. \ref{isp-comp}, Outlier-Aware Loss could improve the PSNR of the output images. A more detailed discussion of $\mathcal{L}_{OA}$ is in Appendix \ref{sec:app-rfl}.

\begin{equation}
  \Delta = I^{SYE} - I^{GT} = \{\delta_{i,j}|i\in[0,H-1] ,j\in[0,W-1]\}
  \label{eq-delta}
\end{equation}

\begin{equation}
  \mathcal{L}_{OA}=\frac{1}{HW}\sum^{H-1}_{i=0}\sum^{W-1}_{j=0} \Big[ {|\delta_{i,j}|}^p\times \Big( 1-e^{-\alpha{|\delta_{i,j}-\mu|}^p/b} \Big) \Big]
  \label{eq-lp}
\end{equation}

\subsection{Revised re-parameterization with enhancement by $1\times 1$ convolution}

All the convolution layers in SYENet shall be re-parameterized as Fig. \ref{sye-rep} for inference. The convolution block in the training phase is expressed as Eq. \ref{eq-rep}.

\setlength{\arraycolsep}{0.0em}
\begin{eqnarray}
I^{(out)} &=& \mathbf{CONV}_{1\times 1}\Big( \mathbf{CAT} \big( \{\mathbf{CONV}_{\Phi}(I^{(in)})| \Phi \} \big) \Big) % \nonumber\\  
%    & & \mathbf{BN} (\mathbf{CONV}_{\Phi}(I^{(in)})) | \Phi \})) %\nonumber\\
%    & & \Phi \in\{1\times 1, 1\times 3, 3\times 1, 3\times 3, 5\times 5 \} \} ))
  \label{eq-rep}
\end{eqnarray}

After the re-parameterization, the complex concatenation of several convolutions, half followed by batch normalization layers, shall be converted back to a single convolution layer as Eq. \ref{eq-rep-inf} for accelerating inference.

\begin{equation}
  I^{(out)} = \mathbf{CONV}_{5\times 5}(I^{(in)})
  \label{eq-rep-inf}
\end{equation}

Compared with the previous re-parameterization techniques, in SYENet, an improvement by one extra convolution layer with the kernel size of $1\times 1$ is implemented after the concatenation to score the importance of each channel. Meanwhile, this structure can be re-parameterized like addition fusion. Compared with RepVGG block \cite{ding2021repvgg}, our revised ConvRep block with $1\times 1$ convolution, which simulates the function of channel attention, could improve the PSNR by 2.1932dB as shown in Table \ref{table-abs}. 

%%%%%%%%%%%%%%%%%%%%%%%%%%%%%%%%%%%%%%%%%%%%%%%%%%%%%%%%%%%%%%%%%%%%%%%%%%%%%%%%%

\subsection{Simple Yet Effective (SYE) Network}

%%%%%%%%%%%%%%%%%%%%%%%%%%%%%%%%%%%%%%%%%%%%%%%%%%%%%%%%%%%%%%%%%%%%%%%%%%%%%%%%%
%%%% SR TABLE

\begin{table*}[htbp]
\centering
\resizebox{2.0\columnwidth}{!}{
    \begin{tabular}{lcccccccccc}
    \toprule
                Method & Scale & \#P & Avg latency(ms) & FPS(2K) & Set5 & Set14 & BSD100 & BSD100 Score & Urban100 & Urban100 Score\\
    \midrule
                CISR\cite{gunawan2022cisrnet} & $\times 2$ & 9.60M & 1K+ & $<$1 & 28.94/0.8160 & 26.78/0.7080 & 26.08/0.6590 & - & 24.93/0.7270 & - \\
                VSDR\cite{kim2016accurate} & $\times 2$ & 0.65M & 1K+ & $<$1 & 37.53/0.9587 & 33.03/0.9124 & 31.90/0.8960 & - & 30.76/0.9140 & - \\
                DBPN\cite{haris2018dbpn} & $\times 2$ & 5.95M & 1K+ & $<$1 & 38.09/0.9600 & 33.85/0.9190 & 32.27/0.9000 & - & 32.55/0.9324 & - \\
                RDN\cite{zhang2018residual} & $\times 2$ & 22.12M & 1K+ & $<$1 & 38.24/0.9614 & 34.01/0.9212 & 32.34/0.9017 & - & 32.89/0.9353 & - \\
                RCAN\cite{zhang2018image} & $\times 2$ & 12.47M & 1K+ & $<$1 & 38.27/0.9614 & 34.12/0.9216 & 32.41/0.9027 & - & 33.34/0.9384 & - \\
                HAN\cite{niu2020han} & $\times 2$ & 64.61M & 1K+ & $<$1 & 38.27/0.9614 & 34.16/0.9217 & 32.41/0.9027 & - & 33.35/0.9385 & - \\
                DRLN\cite{anwar2022drln} & $\times 2$ & 34.43M & 1K+ & $<$1 & 38.27/0.9616 & 34.28/0.9231 & 32.44/0.9028 & - & 33.37/0.9390 & - \\
                IPT\cite{chen2021pre} & $\times 2$ & 64.27M & 1K+ & $<$1 & \textbf{38.37}/- & \textbf{34.43}/- & \textbf{32.48}/- & - & \textbf{33.76}/- & - \\
\hline %%%%%% 
                ESPCN\cite{shi2016real}(D0S3) & $\times 2$ & 0.191K & 6.0 & 166 & 29.76/0.9190 & 28.96/0.8810 & 28.69/0.8650 & 1.737 & 26.38/0.8530 & 0.508 \\
                EDSR\cite{lim2017enhanced} & $\times 2$ & 1.37M & 852.0 & 1 & 38.11/0.9601 & 33.92/0.9195 & 32.32/0.9013 & 1.874 & 32.93/0.9351 & 31.438 \\
                SRCNN\cite{dong2014learning} & $\times 2$ & 19.6K & 168.0 & 5 & 36.66/0.9542 & 32.42/0.9063 & 31.36/0.8879 & 2.512 & 29.50/0.8946 & 1.373 \\
                eSR\cite{michelini2022edgesr}(C6D3S15) & $\times 2$ & 7.13K & 119.0 & 8 & 36.58/0.9530 & 32.38/0.9050 & 31.25/0.8850 & 3.045 & 29.26/0.8910 & 1.389 \\
                SCSRN\cite{ignatov2022efficient} & $\times2$ & 50.0K & 101.0 & 10 & 36.90/0.9565 & 32.59/0.9087 & 31.42/0.8904 & 4.541 & 29.63/0.8992 & 2.734 \\
                FSRCNN\cite{chao2016fsrcnn}(D56S12M4) & $\times 2$ & 15.44K & 87.6 & 11 & 36.74/0.9541 & 32.45/0.9070 & 31.34/0.8870 & 4.686 & 29.42/0.8950 & 2.356 \\
                ABPN\cite{du2021anchorbased} & $\times2$ & 33.5K & 86.6 & 12 & 36.72/0.9556 & 32.49/0.9076 & 31.33/0.8891 & 4.675 & 29.39/0.8955 & 2.286 \\
                HOPN\cite{ignatov2022efficient} & $\times2$ & 32.2K & 61.7 & 16 & 36.27/0.9534 & 32.19/0.9049 & 31.11/0.8865 & 4.836 & 28.90/0.8885 & 1.627 \\
                TPSR-D2\cite{Lee2020JourneyTT} & $\times2$ & 60.8K & 105.0 & 10 & 37.18/0.9578 & 32.84/0.9112 & 31.64/0.8935 & 5.925 & 30.24/0.9073 & 6.126 \\
                FSRCNN\cite{chao2016fsrcnn}(D32S6M4) & $\times 2$ & 5.78K & 48.9 & 20 & 36.29/0.9510 & 32.20/0.9040 & 31.10/0.8840 & 6.018 & 28.91/0.8860 & 2.081 \\
                ESPCN\cite{shi2016real}(D64S32) & $\times 2$ & 24.48K & 54.8 & 18 & 36.64/0.9530 & 32.46/0.9070 & 31.32/0.8870 & 7.286 & 29.37/0.8930 & 3.514 \\
                eSR\cite{michelini2022edgesr}(K3C1) & $\times 2$ & 0.105K & 3.5 & 282 & 33.15/0.9280 & 30.16/0.8820 & 29.66/0.8620 & 11.422 & 26.94/0.8570 & 1.873 \\
                ESPCN\cite{shi2016real}(D22S32) & $\times 2$ & 9.2K & 31.0 & 32 & 36.70/0.9530 & 32.47/0.9070 & 31.35/0.8870 & 13.426 & 29.44/0.8940 & 6.845 \\
                Compiler-Aware NAS\cite{wu2022compiler} & $\times2$ & 11K & 31.6 & 27 & 37.19/0.9572 & 32.80/0.9099 & 31.60/0.8919 & 15.654 & 30.15/0.9054 & 15.100 \\
                FSRCNN\cite{chao2016fsrcnn}(D6S3M1) & $\times 2$ & 1.08K & 8.3 & 121 & 35.36/0.9430 & 31.52/0.8980 & 30.64/0.8780 & 18.740 & 28.01/0.8700 & 3.542 \\
\hline
                SYENet (\textbf{Ours}) & $\times 2$ & 4.932K & 16.5 & 60 & 36.84/0.9564 & 32.62/0.9079 & 31.52/0.8907 & \textbf{31.928} & 30.37/0.9029 & \textbf{46.681} \\
\hline
\hline
                CISR\cite{gunawan2022cisrnet} & $\times 4$ & 9.93M & 1K+ & $<$1 & 25.03/0.7020 & 23.88/0.5960 & 23.83/0.6590 & - & 21.86/0.5820 & - \\
                VSDR\cite{kim2016accurate} & $\times 4$ & 0.65M & 1K+ & $<$1 & 31.35/0.8838 & 28.01/0.7674 & 27.29/0.7261 & - & 25.18/0.7524 & - \\
                RDN\cite{zhang2018residual} & $\times 4$ & 22.27M & 1K+ & $<$1 & 32.47/0.8990 & 28.81/0.7871 & 27.72/0.7419 & - & 26.61/0.8028 & - \\
                RCAN\cite{zhang2018image} & $\times 4$ & 12.61M & 1K+ & $<$1 & 32.63/0.9002 & 28.87/0.7889 & 27.77/0.7436 & - & 26.82/0.8087 & - \\
                HAN\cite{niu2020han} & $\times 4$ & 64.20M & 1K+ & $<$1 & \textbf{32.64}/0.9002 & 28.90/0.7890 & 27.80/0.7442 & - & 26.85/0.8094 & - \\
                DBPN\cite{haris2018dbpn} & $\times 4$ & 10.43M & 1K+ & $<$1 & 32.47/0.8980 & 28.82/0.7860 & 27.72/0.7400 & - & 26.38/0.7946 & - \\
                IPT\cite{chen2021pre} & $\times 4$ & 64.41M & 1K+ & $<$1 & \textbf{32.64}/- & \textbf{29.01}/- & 27.82/- & - & \textbf{27.26}/- & - \\
                DRLN\cite{anwar2022drln} & $\times 4$ & 34.58M & 1K+ & $<$1 & 32.63/0.9002 & 28.94/0.7900 & \textbf{27.83}/0.7444 & - & 26.98/0.8119 & - \\
\hline %%%%%% 
                SRCNN\cite{dong2014learning} & $\times 4$ & 67.6K & 167.0 & 5 & 30.48/0.8628 & 27.49/0.7503 & 26.90/0.7101 & 0.939 & 24.52/0.7221 & 0.990 \\
                EDSR\cite{lim2017enhanced} & $\times 4$ & 1.52M & 418.0 & 2 & 32.46/0.8968 & 28.80/0.7876 & 27.71/0.7420 & 1.153 & 26.64/0.8033 & 7.475 \\
                eSR\cite{michelini2022edgesr}(C8D9S6) & $\times 4$ & 15.0K & 131.0 & 7 & 30.62/0.8060 & 27.48/0.7510 & 26.93/0.7140 & 1.248 & 24.42/0.7180 & 1.099 \\
                ABPN\cite{du2021anchorbased} & $\times 4$ & 62K & 50.1 & 20 & 30.61/0.8684 & 27.61/0.7578 & 26.94/0.7160 & 3.310 & 24.53/0.7275 & 3.347 \\
                SCSRN\cite{ignatov2022efficient} & $\times 4$ & 73.9K & 31.0 & 32 & 30.75/0.8719 & 27.75/0.7616 & 27.02/0.7188 & 5.955 & 24.69/0.7343 & 6.730 \\
                TPSR-D2\cite{Lee2020JourneyTT} & $\times 4$ & 61K & 31.7 & 32 & 30.99/0.8761 & 27.85/0.7639 & 27.08/0.7211 & 6.349 & 24.81/0.7393 & 7.798 \\
                HOPN\cite{ignatov2022efficient} & $\times 4$ & 41.3K & 20.8 & 48 & 30.25/0.8598 & 27.35/0.7515 & 26.80/0.7115 & 6.564 & 24.27/0.7159 & 5.622 \\
                ESPCN\cite{shi2016real}(D64S32) & $\times 4$ & 27.3K & 19.5 & 51 & 30.57/0.8580 & 27.50/0.7520 & 26.92/0.7150 & 8.268 & 24.42/0.7180 & 7.382 \\
                FSRCNN\cite{chao2016fsrcnn}(D32S6M1) & $\times 4$ & 5.78K & 11.8 & 84 & 30.16/0.8450 & 27.19/0.7420 & 26.74/0.7070 & 10.646 & 24.09/0.7020 & 7.21 \\
                ESPCN\cite{shi2016real}(D1S3) & $\times 4$ & 0.541K & 5.65 & 176 & 28.93/0.8200 & 26.49/0.7250 & 26.25/0.6940 & 11.273 & 23.56/0.6800 & 7.803 \\
                FSRCNN\cite{chao2016fsrcnn}(D44S12M4) & $\times 4$ & 13.26K & 13.1 & 76 & 30.61/0.8610 & 27.52/0.7530 & 26.94/0.7160 & 12.654 & 24.44/0.7210 & 11.298 \\
                FSRCNN\cite{chao2016fsrcnn}(D6S1M1) & $\times 4$ & 0.953K & 5.7 & 125 & 29.31/0.8230 & 26.62/0.7300 & 26.41/0.6990 & 13.949 & 23.62/0.6830 & 8.331 \\
                eSR\cite{michelini2022edgesr}(K3C2) & $\times 4$ & 0.844K & 3.68 & 271 & 28.64/0.8060 & 26.12/0.7120 & 26.13/0.6840 & 14.655 & 23.28/0.6680 & 8.011 \\
                ESPCN\cite{shi2016real}(D16S32) & $\times 4$ & 10.48K & 11.3 & 88 & 30.59/0.8590 & 27.53/0.7530 & 26.95/0.7150 & 14.874 & 24.43/0.7190 & 12.918 \\
    \hline
                SYENet (\textbf{Ours}) & $\times 4$ & 5.268K & 9.92 & 100 & 30.33/0.8646 & 27.43/0.7532 & 27.02/0.7214 & \textbf{18.670} & 24.91/0.7299 & \textbf{28.682} \\
    \hline
    \end{tabular}
}
%\vspace{-2mm}
\quad \\
    \caption{Comparison on super-resolution issue between the results by PSNR(dB), SSIM, and comprehensive score with SOTA: The methods are classified into big models with latency larger than 1K(ms) and small models. Big models are ranked by PSNR on BSD100 dataset, and small models are ranked by score in Eq. \ref{eq-score} on BSD100 dataset.}
    \label{table-sr}
    %\vspace{-5mm}
\end{table*}

%%%%%%%%%%%%%%%%%%%%%%%%%%%%%%%%%%%%%%%%%%%%%%%%%%%%%%%%%%%%%%%%%%%%%%%%%%%%%%%%%%%%
%%% LLE TABLE

\begin{table}[htbp]
\centering
\resizebox{1.0\columnwidth}{!}{
    \begin{tabular}{lcccc}
    \toprule
                Method & \#P(M) & Mobile GPU latency(ms) & PSNR & SSIM \\
    \midrule
                ZeroDCE\cite{guo2020ZeroDCE} & 0.08 & 858 & 14.83 & 0.531 \\
                UFormer\cite{wang2022uformer} & 5.29 & - & 16.27 & 0.771 \\ %1000+
                3D-LUT\cite{zeng20223dlut} & 0.60 & - & 16.35 & 0.585 \\
                Kind++\cite{zhang2021kind++} & 8.28 & - & 16.36 & 0.820 \\ %1000+
                LIME\cite{tang2021lime} & - & - & 16.76 & 0.650 \\ %1000+
                RetiNexNet\cite{wei2018deep} & 0.84 & - & 17.90 & 0.562 \\ %1000+
                DRBN\cite{yang2020drbn} & 0.58 & - & 19.55 & 0.746 \\
                MBLLEN\cite{Lv2018MBLLEN} & 20.47 & - & 20.86 & 0.702 \\ %1000+
                KIND\cite{zhang2019kind} & 8.16 & - & 21.30 & 0.790 \\ %1000+
                Night Enhancement\cite{jin2022unsupervised} & 40.39 & - & 21.52 & 0.765 \\ %1000+
                IPT\cite{chen2021pre} & 115.63 & - & 22.67 & 0.504 \\ %1000+
                IAT\cite{cui2022iat} & 0.09 & 668 & 23.38 & 0.809 \\
                RCT\cite{kim2021rct} & - & - & 23.43 & 0.788 \\
                MIRNet\cite{hyewon2020mirnet} & - & - & 24.14 & 0.830 \\
                HWMNet\cite{fan2022hwmnet} & 66.56 & - & 24.14 & \textbf{0.930} \\ %1000+
                MAXIM\cite{tu2022maxim} & 14.14 & - & 24.24 & 0.863 \\ %1000+
                LLFlow\cite{wang2022low} & 17.42 & - & \textbf{25.19} & 0.850 \\ %1000+
    \hline
                SYENet (\textbf{Ours}) & \textbf{0.005} & 33.4 & 22.59 & 0.807 \\
    \hline
    \end{tabular}
}
%\vspace{-3mm}
\quad \\
    \caption{Comparison on low-light enhancement issue between the results by PSNR(dB) and SSIM with SOTA: The '-' mark in the Mobile GPU latency column refers that the latency of that model is larger than 1000ms.}
    \label{table-lle}
% \vspace{-3mm}
\end{table}

%%%%%%%%%%%%%%%%%%%%%%%%%%%%%%%%%%%%%%%%%%%%%%%%%%%%%%%%%%%%%%%%%%%%%%%%%%%%%%%%%%%%%
%%% LLE SR

\begin{figure*}[t]
  \centering
    \includegraphics[width=1.0\linewidth]{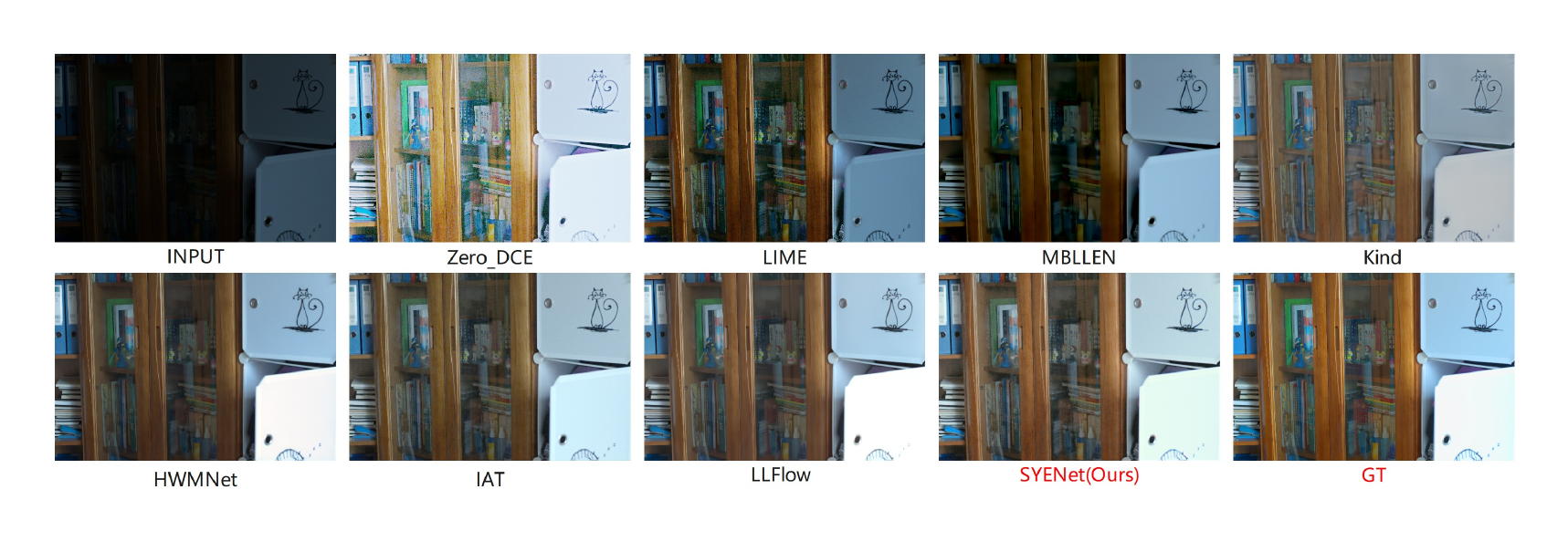}
    %\vspace{-14mm}
   \caption{Low-light enhancement Comparison: The results reveal that our method could competitively recover the illuminance information. More comparisons of qualitative results are presented in the Appendix \ref{apdx:lol-comps}.}
   \label{lle-comp}
% \vspace{-5mm}
\end{figure*}

\begin{figure*}[t]
  \centering
    \includegraphics[width=1.0\linewidth]{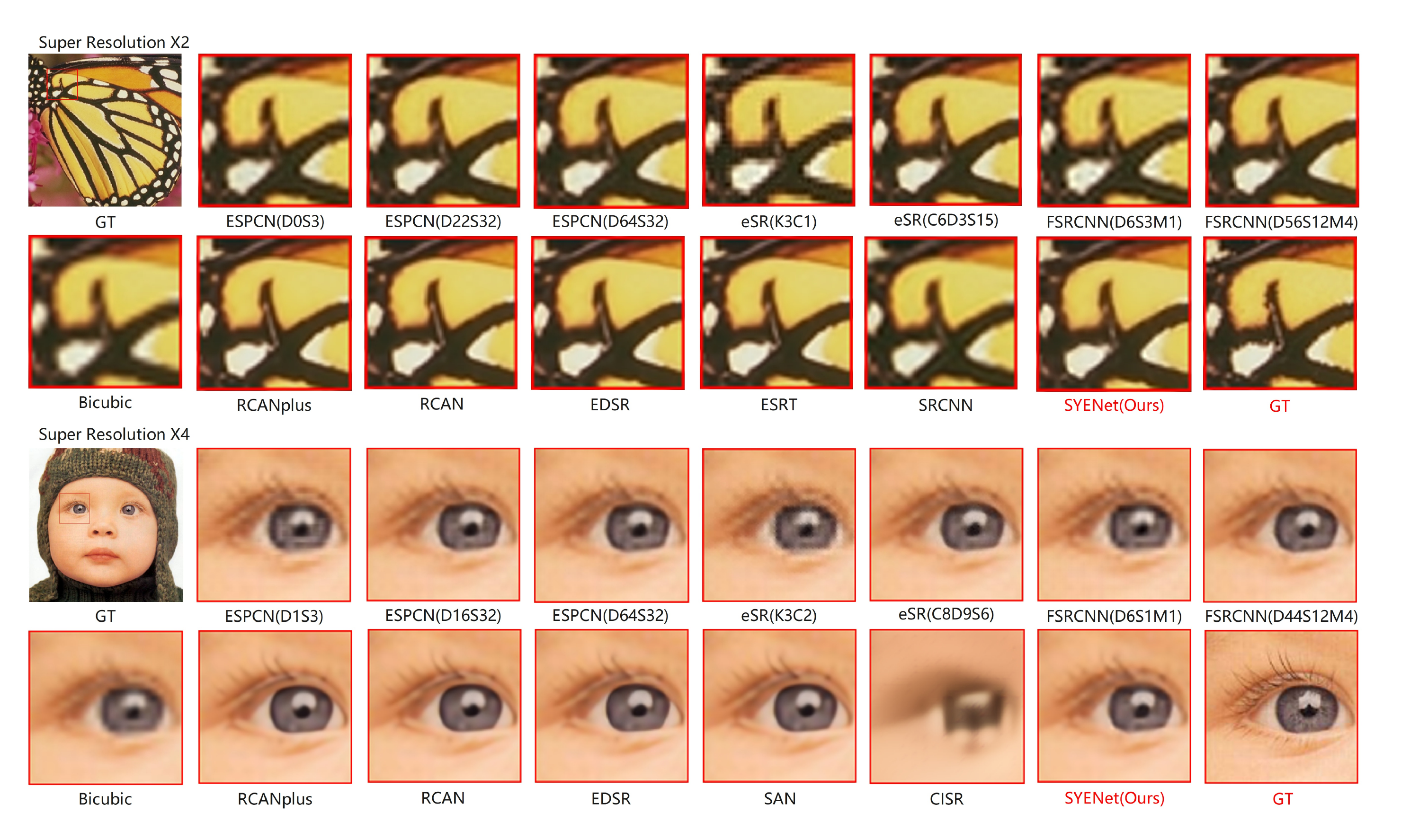}
    %\vspace{-5mm}
   \caption{$\times$2 and $\times$4 SR comparisons with SOTA models: It is observed that our efficient model could generate output images with a similar quality compared with other large models. It is recommended to zoom in to observe the details.} 
   \label{sr-comp}
%\vspace{-5mm}
\end{figure*}

%%%%%%%%%%%%%%%%%%%%%%%%%%%%%%%%%%%%%%%%%%%%%%%%%%%%%%%%%%%%%%%%%%%%%%%%%%%%%%%%%%%%%

\begin{table}[htbp]
\centering
\resizebox{1.0\columnwidth}{!}{
    \begin{tabular}{lccccccccc}
    \hline
                Method & Model Size(MB) & PSNR & SSIM & GPU Runtime(ms) & Score \\
\hline\hline
                DANN-ISP & 29.4 & 23.10 & 0.8648 & 583 & 0.13 \\
                MiAlgo & 117 & 23.65 & 0.8673 & 1164 & 0.14 \\
                CASIA 1st & 205 & \textbf{24.09} & \textbf{0.8840} & 1044 & 0.28 \\
                rainbow & 1.0 & 21.66 & 0.8399 & 28 & 0.36 \\
                JMU-CVLab & 0.041 & 23.22 & 0.8281 & 182 & 0.48 \\
                HITZST01 & 1.2 & \textbf{24.09} & 0.8667 & 482 & 0.60 \\
                ENERZAi & 4.5 & 24.08 & 0.8778 & 212 & 1.35 \\
                MINCHO & 0.067 & 23.65 & 0.8658 & 41.5 & 3.80 \\
                HITZST01 & 0.060 & 23.89 & 0.8666 & 34.3 & 6.41 \\
                ENERZAi & 0.077 & 23.8 & 0.8652 & 18.9 & 10.27 \\
                MiAlgo & 0.014 & 23.33 & 0.8516 & \textbf{6.8} & 14.87 \\
    \hline
                SYENet(\textbf{Ours}) & 0.029 & 23.96 & 0.8543 & 11.4 & \textbf{21.24} \\
    \hline
    \end{tabular}
}
%\vspace{-3mm}
\quad \\
    \caption{Comparison on ISP performance by PSNR(dB) and SSIM with algorithms of MAI2022 ISP Challenge\cite{ignatov2022maiisp}: even though the PSNR of our method is not the highest, the comprehensive performance of our method measured by score (Eq. \ref{eq-score}) is the highest.}
    \label{table-isp}
%\vspace{-2mm}
\end{table}

% \subsection{SYENet architecture}

The SYENet consists of 5 parts: head block, the first and second asymmetrical block, channel attention block, and tail block, which are assigned as $\mathbf{H}$, $\mathbf{A_1}$, $\mathbf{A_2}$, $\mathbf{CA}$ and $\mathbf{T}$. The head block is arranged for the preference of different tasks. The asymmetrical blocks are utilized to generate texture features and pattern information, which afterward shall be fused using multiplication. With the network input as $I^{(in)}$, the output of the first asymmetrical block $I^{(a_1)}$ and second $I^{(a_2)}$ are expressed as below, in which the subscript $(c)$ and $(s)$ represent the complex and the simple asymmetric branch respectively. 

\begin{equation}
  % I^{(a_1)} = \mathbf{A}^{(c)}_1(\mathbf{H}(I^{(in)})) \odot \mathbf{A}^{(s)}_1(\mathbf{H}(I^{(in)})) + B_1
  I^{(a_1)} = \mathbf{QCU}\Big( \mathbf{A}^{(c)}_1 \big( \mathbf{H}(I^{(in)}) \big), \mathbf{A}^{(s)}_1 \big( \mathbf{H}(I^{(in)}) \big) \Big)
  \label{eq-a1}
\end{equation}

\begin{equation}
  % I^{(a_2)} = \mathbf{A}^{(c)}_2(I^{(a_1)}) \odot \mathbf{A}^{(s)}_2(I^{(a_1)}) + B_2
  I^{(a_2)} = \mathbf{QCU}\Big( \mathbf{A}^{(c)}_2(I^{(a_1)}), \mathbf{A}^{(s)}_2(I^{(a_1)}) \Big)
  \label{eq-a2}
\end{equation}

The squeeze-and-excitation block is adopted and employed as the channel attention block, enhancing the expressiveness using global information to compensate for the disadvantage of the small receptive field. Hence the output of SYENet is expressed as Eq. \ref{eq-out}, in which $\otimes$ is channel-wise multiplication.

% in which the global pooling, full-connected layer, and ReLU activation are replaced by adaptive average pooling, convolutional layer, and PReLU respectively as shown in Fig. \ref{ca}. 

\begin{equation}
  I^{(out)}=\mathbf{T}\Big( \mathbf{CONV}\big( \mathbf{CA}(I^{(a_2)}) \otimes I^{(a_2)}\big) \Big)
  \label{eq-out}
\end{equation}

%---------------------------------------------------------------------------------------------------------
%------------------------------------------------------------------------

\begin{table}[htbp]
\centering
\resizebox{1.0\columnwidth}{!}{
    \begin{tabular}{ccccc|cc}
    \hline
                $\mathcal{L}_{OA}$ & ConvRep & CA & QCU & Two-branch & PSNR & $\Delta$PSNR \\
    \hline\hline
                $\mathcal{L}_1$ & $\surd$ & $\surd$ & $\surd$ & $\surd$ & 24.7200 & +0.1532 \\
                $\surd$ & RepVGGBlock\cite{ding2021repvgg} & $\surd$ & $\surd$ & $\surd$ & 22.6797 & +2.1932 \\
                $\surd$ & $\times$ & $\surd$ & $\surd$ & $\surd$ & 24.6778 & +0.1954 \\
                $\surd$ & $\surd$ & $\times$ & $\surd$ & $\surd$ & 24.0936 & +0.7796 \\
                $\surd$ & $\surd$ & $\surd$ & ADD & $\surd$ & 24.5252 & +0.3480 \\
                $\surd$ & $\surd$ & $\surd$ & CAT+CONV & $\surd$ & 24.5427 & +0.3305 \\
                $\surd$ & $\surd$ & $\surd$ & MUL & $\surd$ & 24.7971 & +0.0761 \\
                $\surd$ & $\surd$ & $\surd$ & $\surd$ & $\times$ & 24.5510 & +0.3222 \\
    \hline
                $\surd$ & $\surd$ & $\surd$ & $\surd$ & $\surd$ & 24.8732 & - \\
    \hline
    \end{tabular}
}
\quad \\
%\vspace{-3mm}
    \caption{Ablation study towards $\mathcal{L}_{OA}$(Outlier-Aware Loss) by $\mathcal{L}_1$(L1 loss), our re-parameterized convolution(ConvRep) by RepVGGBlock\cite{ding2021repvgg}, CA(channel attention) by no CA, QCU(Quadratic Connection Unit) feature fusion by ADD(element-wise addition), MUL(element-wise multiplication), and CAT+CONV(concatenation followed by convolution) feature fusion, and two-branch asymmetric re-parameterized block by single branch re-parameterized block. The ablation study is conducted on ISP task.}
    \label{table-abs}
% \vspace{-3mm}
\end{table}
% $\surd$ & $\surd$ & SA & ADD & $\surd$ & 24.3758 & +0.3711 \\

\begin{table}[htbp]
% \vspace{-3mm}
\centering
\resizebox{1.0\columnwidth}{!}{
    \begin{tabular}{c|cccc}
    \hline
        Models & \multicolumn{4}{c}{SYENet(ISP)} \\
    \hline
        Metric & PSNR $\uparrow$ & LPIPS $\downarrow$ & FID $\downarrow$ & KID $\downarrow$ \\
        % & \multicolumn{4}{|c|}{SYENet} & & \\ 
    \hline
        L1 Loss $\mathcal{L}_1$ & 24.7200 & 0.1681 & 28.0420 & 0.0095 \\
        Outlier-Aware Loss(Ours) $\mathcal{L}_{OA}$ & \textbf{24.8732} & \textbf{0.1664} & \textbf{27.2182} & \textbf{0.0086} \\
    \hline
    \end{tabular}
}
\quad \\
%\vspace{-3mm}
    \caption{The performance of SYENet trained by two loss functions measured by different metrics: Outlier-Aware Loss improves PSNR as well as visual quality reflected by LPIPS, FID, and KID.}
    \label{table-pl}
%\vspace{-1mm}
\end{table}

%------------------------------------------------------------------------
\section{Experiments}
\label{sec:experiments}

The experiments include sophisticated comparisons between SOTA methods with SYENet in (a)ISP, (b)SR, and (c)LLE issues and ablation studies. The evaluation metrics include PSNR and SSIM, but in order to assess the comprehensive performance of models considering both the image quality and efficiency, the comprehensive score Eq. \ref{eq-score} by MAI Challenge \cite{mai2021isp} is introduced, in which constant $C$ is employed for normalization.

\begin{equation}
  \mathrm{Score} = 2^{2\times \mathrm{PSNR}} / (C\times \mathrm{Latency})
  \label{eq-score}
\end{equation}

%\subsection{Evaluation metric}

\subsection{Implementation details}

\textbf{Training Setting.} For \textbf{SR} task, the inputs are $128\times128$ patches with random augmentation of flips and rotations. The Adam optimizer with $\beta_1=0.9$ and $\beta_2=0.999$ and cosine annealing decay policy are utilized. Moreover, for \textbf{ISP} task, the input is preprocessed as Bayer Pattern with $256\times256$ resolution. Before the official training, an MAE-like \cite{he2022masked} unsupervised warming-up phase is deployed to upgrade robustness as described in Appendix \ref{sec:app-warm}. The \textbf{LLE} task follows the settings of the SR task except for the LoL\cite{wei2018deep} dataset.

\textbf{Inference Setting.} We use the Qualcomm Snapdragon 8 Gen 1 mobile SoC as our target runtime evaluation platform. The application we use to test the model runtime is AI benchmark\cite{ignatov2018ai,ignatov2019ai}, which allows to load any custom TensorFlow Lite model\cite{lee2019device} and run it on any Android device with all supported acceleration options. In our approach, we transform our Pytroch model into tflite model.

\textbf{Datasets.} The dataset for \textbf{ISP} task is MAI21\cite{Ignatov2021maireport} adjusted using conversion by classical algorithm and warping by PDC-Net \cite{truong2021learning}. For the \textbf{SR} task, we use the DIV2K\cite{timofte2017ntire} for training and set5\cite{bevilacqua2012low}, set14\cite{zeyde2010single}, BSD100\cite{martin2001database}, and Urban100\cite{huang2015single} for testing. For the \textbf{LLE} task, we use LoL\cite{wei2018deep}. % dataset.

\subsection{Comparison with SOTA}

In this study, we compare our proposed model with a variety of SOTA methods, from models with extreme complexity and distinct image quality to lightweight models with excellent efficiency and reasonably good output quality.

% Fig. \ref{ivq-sr2} 
\textbf{Super Resolution.} As illustrated in Fig. \ref{ivq-sr4}, and Table \ref{table-sr}, SYENet achieves a competitive performance, which is roughly only 2dB lower than the highest PSNR but with only $0.17\%$ of its parameters, as well as x100 times faster for inference. SYENet outperforms other lightweight models by 1 to 7dB, and as indicated by Table \ref{table-sr}, SYENet gets far better scores than other lightweight models. The comparison between images by SYENet and other SOTA models with scale factors of $\times 2$ and $\times 4$ is shown in Fig. \ref{sr-comp}.

\textbf{Low-light Enhancement.} The enhanced low-light images obtained by a variety of models are shown in Fig. \ref{lle-comp}, and it is indicated that the images by SYENet could almost reach the objective quality of those by SOTA methods. More photos for comparison can be found in Appendix \ref{apdx:lol-comps}. Finally, the objective measurements of SOTA algorithms and SYENet are shown in Table \ref{table-lle}, which refers that SYENet achieves a competitive image quality at a rather faster speed using roughly only 0.01\% of the size by the latest SOTA models.

\textbf{Image Signal Processing.} The comparison of performance and comprehensive scores by SYENet and the algorithms of MAI ISP Challenge participants is shown in Fig. \ref{ivq-isp} and Table \ref{table-isp}. It is indicated that the comprehensive score by SYENet is significantly higher than the challenge-winning algorithm.

\subsection{Ablation study}

% The ablation study, in which the degradation of Outlier-Aware Loss $\mathcal{L}_{OA}$, ConvRep block as Fig. \ref{sye-rep}, and QCU are set to be L1 loss $\mathcal{L}_1$, RepVGGBlock\cite{ding2019acnet}, and three fusion methods of element-wise addition, concatenation plus convolution, and element-wise multiplication, respectively shows that those components indeed improve the PSNR. In addition, Table\ref{table-pl} shows that $\mathcal{L}_{OA}$ could consistently improve the visual quality measured by LPIPS \cite{zhang2018lpips}, FID, and KID.

In the ablation study, the Outlier-Aware Loss $\mathcal{L}_{OA}$, ConvRep block as Fig. \ref{sye-rep}, channel attention, QCU, and asymmetric branch block are degraded to be L1 loss $\mathcal{L}_1$, RepVGGBlock\cite{ding2019acnet}, no channel attention, three fusion methods (element-wise addition, concatenation plus convolution, and element-wise multiplication), and single branch block respectively. It shows that those components or methods indeed improve the PSNR. In addition, $\mathcal{L}_{OA}$ could improve the visual quality as Table \ref{table-pl}.

%------------------------------------------------------------------------
\section{Conclusion and Future Work}
\label{sec:conclusion}
In this paper, we proposed SYENet, a novel and end-to-end mobile network for multiple low-level vision tasks with two asymmetric branches, \textbf{QCU}, revised re-parameter convolution, and channel attention. We also developed the \textbf{Outlier-Aware Loss} for better training. With these simple yet effective methods, SYENet is able to achieve 2K60FPS real-time performance on mobile devices for ISP, SR, and LLE tasks with the best visual quality. 

While these initial results are promising, many challenges still remain. The most critical one is that the proposed network cannot handle all the low-level vision tasks, such as denoise and video SR. There's still room to improve the run-time efficiency by better utilization of limited hardware resources. In the future, we will focus on a more universal network architecture with reduced computation complexity.

{\small
\bibliographystyle{ieee_fullname}
\bibliography{egbib}
}

\clearpage \appendix

\onecolumn

\label{sec:appendix}

\section{Outlier-Aware Loss}
\label{sec:app-rfl}

\subsection{Analysis of Outlier-Aware Loss}

We produced some histograms of the pixel-to-pixel difference values obtained by various models on several datasets, and find that often the difference values concentrate on the mean that is close to zero and show a similar frequency curve as some bell-shaped distribution with a small variance. Thus, we postulate that the true value is equal to the predicted value plus error as $y = \hat{y} + \delta$, in which $\delta$ follows an unknown distribution with a mean of 0, which is not always the case and influenced by the data distribution, loss function, and the model itself. Sometimes, the results generated by some models may have a very significant bias. But in this case, with a similar frequency curve as the Laplacian distribution due to the $\mathcal{L}_1$ loss, the losses by properly predicted pixels form the majority of the total loss when the network converges. Therefore, decreasing the losses in the majority well predicted pixels could rapidly reach a smaller total loss while ignoring the outliers. However, we think that the outliers or the badly predicted pixels as the minority outside the confidence interval shall be the focus of the optimization. Therefore, inspired by Focal Loss\cite{lin2017focalloss} which focuses more on the hard examples, we propose Outlier-Aware Loss $\mathcal{L}_{OA}$. 

We apply the idea of Focal Loss to image restoration problems. Following that Focal Loss optimizes cross entropy, we decide to optimize $\mathcal{L}_1$ loss. Compared to $\mathcal{L}_1=\sum_i|y_i-\hat{y}_i|$, $\mathcal{L}_2=\sum_i{(y_i-\hat{y}_i)^2}$ is more sensitive to outliers as the difference is squared, which causes the suppression of high-frequency details involving the regression-to-the-mean issue \cite{gondal2018ECCVWUET}. Since $\mathcal{L}_2$ loss basically causes the predicted images to be blurry\cite{Mathieu2015DeepMV}, $\mathcal{L}_1$ loss is widely utilised in image restoration networks. Thus, we decide to figure out a new loss function for the image restoration problem which shall be in the form of $\mathcal{L}_1 \times W$ where $W$ represents the weight indicating the degree of hardness on properly predicting this pixel. 

$\mathcal{L}_2$ loss assumes that data is drawn from Gaussian distribution\cite{Mathieu2015DeepMV}, and similarly, $\mathcal{L}_1$ loss comes from the assumption that the data is drawn from Laplacian distribution as shown in Eq. \ref{eq-laplacian} where $\mu$ is the mean and $b$ is the scale parameter. In this study, we decide to apply the Laplacian distribution as the weight decay curve to rearrange the importance of the loss generated by different pixels. The loss of the terribly predicted pixels will be enlarged.

\begin{equation}
  f(x|\mu, b) = \frac{1}{2b}\mathrm{exp}(-|x-\mu|/b)
  \label{eq-laplacian}
\end{equation}

\begin{figure*}
  \centering
  \begin{subfigure}{1.0\linewidth}
    \centering
    \includegraphics[width=1.0\linewidth]{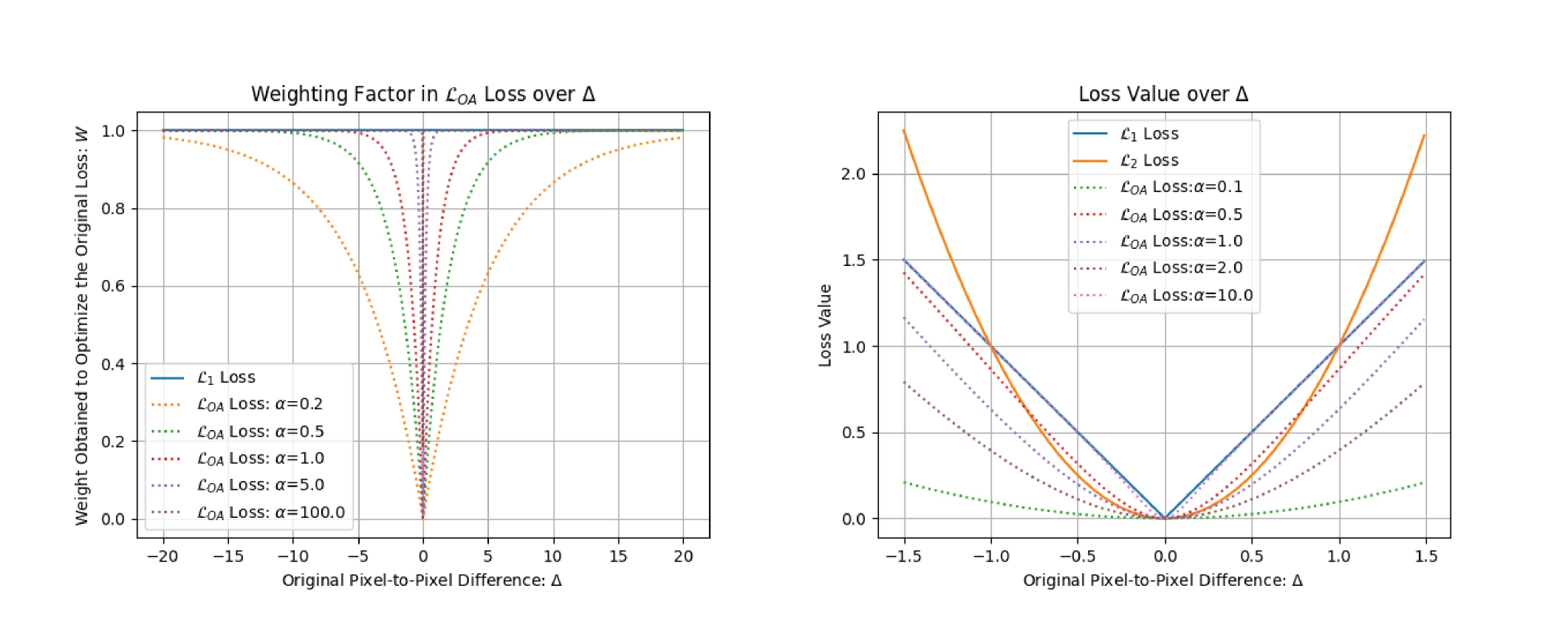}
    \caption{\textbf{(left)}Weight $W(x|\mu, b) = 1-\mathrm{exp}(-\alpha{|x-\mu|}/b)$ for adjusting the original loss over the difference between the network output and ground truth $x=\Delta=I^{(Output)}-I^{(GT)}$ as the tunable hyperparameter $\alpha$ varies. For simplicity, it is assumed that the mean $\mu$ and the scale parameter $b$ of $\Delta$ are 0 and 1 respectively. It shows that larger $\alpha$ makes $\mathcal{L}_{OA}$ approach $\mathcal{L}_1$, while smaller $\alpha$ reduces the weights of the loss by well predicted pixels to a larger degree. \textbf{(right)}The loss value over $\Delta$: compared with $\mathcal{L}_2$, $\mathcal{L}_{OA}$ reduces the weights of the well predicted pixels instead of squaring the loss values of badly predicted pixels. Since $\mathcal{L}_{OA} \leq \mathcal{L}_1$, $\mathcal{L}_1$ becomes the asymptote of $\mathcal{L}_{OA}$ as $\alpha$ becomes larger.}
    \label{rfl-wdis}
  \end{subfigure}
  %\vspace{-10mm}
  \hfill
  \centering
  
  \begin{subfigure}{1.0\linewidth}
    \centering
    \includegraphics[width=0.95\linewidth]{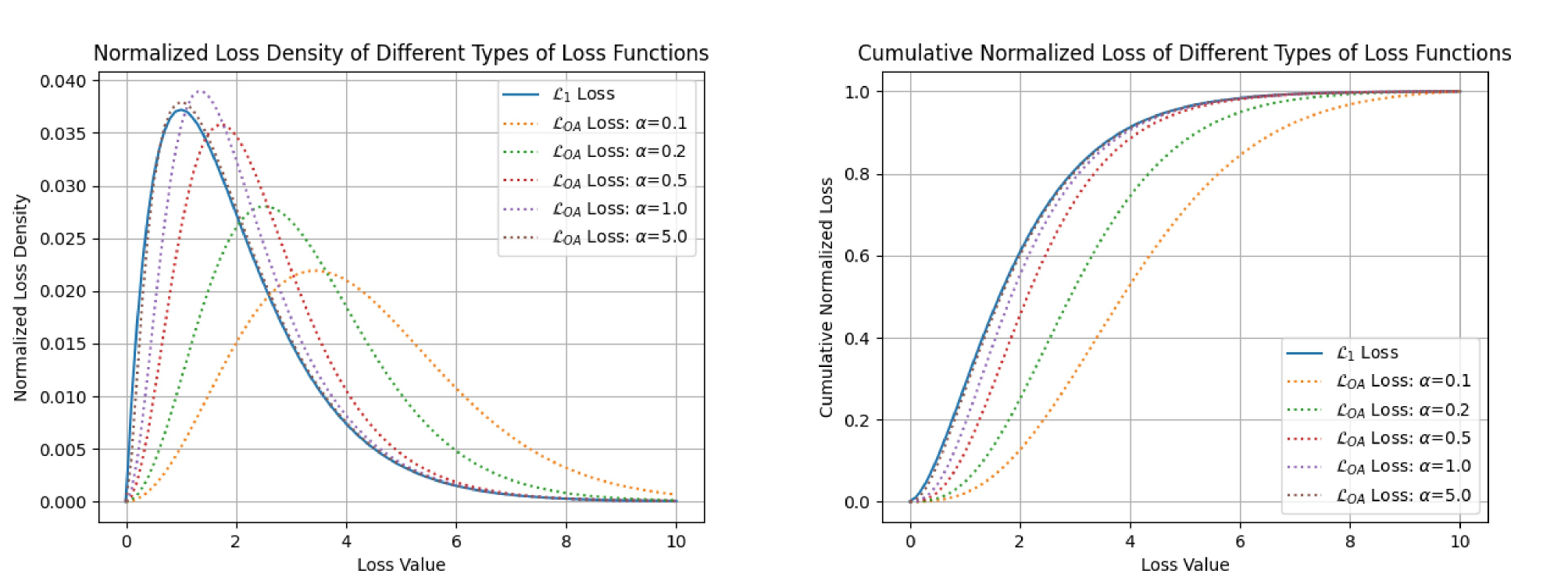}
    \caption{\textbf{(left)}The normalized loss density that we define as loss value times the probability density of that loss value $\mathcal{L}\times p(\mathcal{L})$, which leads to the expectation of loss to be 1 as $\int_{\mathcal{L}}\mathcal{L}\times p(\mathcal{L})d\mathcal{L}=1$. We could understand the loss density as the contribution to the total loss by this loss value. It could be observed that $\mathcal{L}_{OA}$ reshapes the loss density by increasing the proportion of outliers and reducing the proportion of well predicted pixels in the total loss. Specifically, a smaller $\alpha$ causes a larger proportion transfer towards the outliers. Therefore, we successfully increase the importance of badly predicted outliers and decrease the importance of well predicted pixels in the training. Initially, we try to collect the loss values by testing the pre-trained model but found the vibration of the values cause the curve to be too messy. Thus we decide to randomly generate pixel-to-pixel loss values by sampling from the Laplacian distribution since the loss values collected from the test look very close to it. Note that this is not always the case due to the variation of the dataset and network itself. \textbf{(right)}The cumulative normalized loss, which is the integration of normalized loss density, shows the growth of total loss when we integrate the total loss from small loss values to large ones. It is observed that for $\mathcal{L}_1$ loss, the loss values smaller than 3 forms 80\% of the total loss, but for $\mathcal{L}_{OA}$ loss with $\alpha=0.1$, the loss values smaller than 3 only forms 30\% of the total loss.}
    \label{rfl-alphav}
  \end{subfigure}

  %\vspace{-7mm}
  \caption{Analysis of the Outlier-Aware Loss: Combining (a) and (b) could demonstrate the function of $\mathcal{L}_{OA}$ and $\alpha$. It could be seen from all the four figures that larger $\alpha$ make $\mathcal{L}_{OA}$ approach $\mathcal{L}_1$ but would not exceed $\mathcal{L}_1$. So that, we could avoid the regression to the mean problem by $\mathcal{L}_2$.}
  \label{app-rfl}
%\vspace{-7mm}
\end{figure*}

The weight is adjusted to ensure $W\in [0,1]$ and is written as $W(x|\mu, b) = 1-\mathrm{exp}(-\alpha{|x-\mu|}/b)$. And consequently, Outlier-Aware Loss $\mathcal{L}_{OA}$ is written as Eq. \ref{eq-lp2}. The parameter $\alpha$ could be utilised to control the degree of focus we would like to put on the accurately predicted pixels. The smaller $\alpha$ is, the less attention is paid to accurately predicted pixels. $\mathcal{L}_{OA}$ shall be approaching $\mathcal{L}_1$ as $\alpha$ gradually becomes larger, which is shown in Fig. \ref{app-rfl}. It could be observed that the number of the badly predicted outliers is reduced by $\mathcal{L}_{OA}$.

\begin{equation}
  \mathcal{L}_{OA}=\frac{1}{HW}\sum^{H-1}_{i=0}\sum^{W-1}_{j=0} \Big[ {|\delta_{i,j}|}\times \Big( 1-e^{-\alpha{|\delta_{i,j}-\mu|}/b} \Big) \Big]
    \label{eq-lp2}
\end{equation}

%%%%%%%%%%%%%%%%%%%%%%%%%%%%%%%%%%%%%%%%%%%%%%%%%%%%%%%%%%%%%%%%%%%%%%%%%%%%%%%%

The generalizability of $\mathcal{L}_{OA}$ is tested on networks as shown by Table \ref{table-plgen}. It is observed that $\mathcal{L}_{OA}$ could consistently improve the performance of those models.

\begin{table}[htbp]
% \vspace{-3mm}
\centering
\resizebox{1.0\columnwidth}{!}{
    \begin{tabular}{c|cccc|ccc|ccc}
    \hline
        Tasks & \multicolumn{4}{|c}{SR} & \multicolumn{3}{|c}{ISP} & \multicolumn{3}{|c}{LLE}\\
    \hline
        Models & EDSR$\times 4$\cite{lim2017enhanced} & EdgeSR$\times 2$\cite{michelini2022edgesr} & FSRCNN$\times 2$\cite{chao2016fsrcnn} & SYENet(SR)$\times 2$ & MiAigo\cite{ignatov2022maiisp} & JMU\cite{ignatov2022maiisp} & SYENet(ISP) & IAN\cite{ian2021guo} & LUM\cite{zhang2021unsupervised} & SYENet(LLE) \\
        % & \multicolumn{4}{|c|}{SYENet} & & \\ 
    \hline
        $\mathcal{L}_1$  & 28.9164 & 32.0771 & 31.8186 & 33.1691 & 23.5752 & 23.7805 & 24.7200 & 19.8092 & 19.2398 & 20.9717 \\
        $\mathcal{L}_{OA}$(Ours) & \textbf{28.9530} & \textbf{32.1961} & \textbf{32.1109} & \textbf{33.1837} & \textbf{23.8255} & \textbf{23.9428} & \textbf{24.8732} & \textbf{20.2487} & \textbf{19.8516} & \textbf{22.5900} \\ %21.2034
    \hline
    \end{tabular}
}
\quad \\
% \vspace{-3mm}
    \caption{The performance (PSNR) of different models could be consistently improved by Outlier-Aware Loss.}
    \label{table-plgen}
%\vspace{-5mm}
\end{table}

\subsection{Generalized Outlier-Aware Loss}

Apart from image restoration problems where $\mathcal{L}_1$ loss is mostly utilised, the Outlier-Aware Loss $\mathcal{L}_{OA}$ could be further generalized by introducing a norm parameter $p$ and changing the definition of scale parameter $b$ for the consistency between \textbf{MLE}(maximum likelihood estimation) and minimizing the loss function. For example, for problems where $\mathcal{L}_2$ loss outperforms $\mathcal{L}_1$ loss, we could set $p$ to be 2 and define $b$ to be $2\sigma^2$ for the consistency between minimizing $\mathcal{L}_2$ loss and MLE on Gaussian distribution.

\begin{equation}
  \mathcal{L}_{OA}=\frac{1}{HW}\sum^{H-1}_{i=0}\sum^{W-1}_{j=0} \Big[ {|\delta_{i,j}|}^p\times \Big( 1-e^{-\alpha{|\delta_{i,j}-\mu|}^p/b} \Big) \Big]
    \label{eq-lpg}
\end{equation}

As mentioned above, different loss functions imply the MLE of different data distributions. And hence, a variety of loss functions should be proposed and studied to respond to the various data distributions in the real world.

\section{ISP unsupervised warming up strategy}
\label{sec:app-warm}

Warming up strategy used in deep learning usually aims at overcoming optimization challenges early in training\cite{goyal2017accurate}.In our ISP task, besides this aim, we want our network to have the ability to recover the defective pixel, which often occurs in ISP raw data because of the physical issues on the camera sensor, by their neighbor pixels. Inspired by \cite{he2022masked}, we randomly mask $\frac{1}{3}$ of the input by $3\times3$ mask tokens and train the network to recover them with $1e-6$ learning rate for 10 epochs. 

\section{Re-parameterizing the ConvRep block}
\label{sec:app-rep}

The input feature and output feature of ConvRep block in SYENet shall have $C^{(\mathrm{in})}$ and $C^{(\mathrm{out})}$ channels. ConvRep block has $N$ branches, each of which can be designed specifically. Each branch could convert the channel of the input feature to $R\times C^{(\mathrm{out})}$ where $R$ is the hyperparameter. The features by $N$ branches shall be concatenated in channel dimension, and hence the number of channels becomes $N\times R\times C^{(\mathrm{out})}$. Then the final $1\times1$ convolution converts the feature with $N\times R\times C^{(\mathrm{out})}$ channels to output feature with $C^{(\mathrm{out})}$ channels. The reparameterization shall turn the complex ConvRep block back to a normal convolution block for inference.

The convolutional block injected by an input feature tensor $I^{(in)}\in \mathbb{R}^{N\times C^{(in)}\times W\times H}$ shall have a weight matrix $W\in \mathbb{R}^{C^{(out)}\times C^{(in)} \times k_W \times k_H}$ and a bias matrix $B\in \mathbb{R}^{C^{(out)}}$. In SYENet, it is set that all the convolutions generate the output with the same shape of the input feature map by arranging the padding carefully for the purpose of concatenation afterward. Batch normalization could be expressed as Eq. \ref{eq-bn} converting input $X$ to $Y$, in which $\gamma$ and $\beta$ are normally assigned to be $1$ and $0$ respectively for no linear transformation.

\setlength{\arraycolsep}{0.0em}
\begin{eqnarray}
  Y &=& \frac{{X}-\mathbf{E}[X]}{\sqrt{\mathbf{VAR}[X]+\epsilon}}\times\gamma + \beta \nonumber \\
  &=& (\frac{X}{\sqrt{\mathbf{VAR}[X]+\epsilon}} + \frac{-\mathbf{E}[X]}{\sqrt{\mathbf{VAR}[X]+\epsilon}}) \times\gamma + \beta  \nonumber\\ 
  &=& (W^{(\mathrm{bn})}\times X + B^{(\mathrm{bn})})\times\gamma + \beta
  \label{eq-bn}
\end{eqnarray}
\setlength{\arraycolsep}{5pt}

It is obvious that the batch normalization towards the output of the convolution layer could be converted to the batch normalization towards the weight and bias of the convolution layer, as shown in Eq. \ref{eq-ibn1}. In addition, the concatenation of convolution output could also be re-parameterized as the concatenation of the weights and bias of all the convolution layers, as shown by Eq. \ref{eq-convrep}. In the following equations, $\bullet$ means convolution operation. It is assumed the convolution in each branch has $C^{(out)}$ output channels, and hence the concatenation of convolutions from all the branches can be reparameterized as a convolution with $N\times C^{(out)}$ output channels.

\begin{equation}
  F^{(\mathrm{conv})} = \mathbf{CONV}(F^{(\mathrm{in})}) = F^{(\mathrm{in})} \bullet W^{(\mathrm{c})}+B^{(\mathrm{c})}
  \label{eq-conv}
\end{equation}

Substituting Eq. \ref{eq-conv} into the Eq. \ref{eq-bn} could obtain the complete expression of convolution followed by batch normalization, which is shown 

\setlength{\arraycolsep}{0.0em}
\begin{eqnarray}
  F^{(\mathrm{conv+bn})} &=& \mathbf{BN}(F^{(\mathrm{conv})}) \nonumber\\ 
  &=&  \big( W^{(\mathrm{bn})}\times F^{(\mathrm{conv})} + B^{(\mathrm{bn})} \big)\times\gamma + \beta \nonumber\\ 
  &=& \big[ W^{(\mathrm{bn})}(F^{(\mathrm{in})} \bullet W^{(\mathrm{c})}+B^{(\mathrm{c})}) + B^{(\mathrm{bn})} \big] \times\gamma + \beta  \nonumber\\ 
  &=& ( F^{(\mathrm{in})} \bullet W^{(\mathrm{bn})}W^{(\mathrm{c})} + W^{(\mathrm{bn})}B^{(\mathrm{c})} + B^{(\mathrm{bn})} )\times\gamma + \beta \nonumber\\ 
  &=& F^{(\mathrm{in})} \bullet \gamma W^{(\mathrm{bn})}W^{(\mathrm{c})} + \gamma (W^{(\mathrm{bn})}B^{(\mathrm{c})} + B^{(\mathrm{bn})}) + \beta \nonumber\\
  &=& F^{(\mathrm{in})} \bullet W^{(\mathrm{c+bn})} + B^{(\mathrm{c+bn})}
  \label{eq-ibn1}
\end{eqnarray}
\setlength{\arraycolsep}{5pt}

%\setlength{\arraycolsep}{0.0em}
%\begin{eqnarray}
%  F^{(\mathrm{conv+bn})} &=& K_1W^{(\mathrm{bn})}\times (F^{(\mathrm{in})} \bullet W^{(\mathrm{c})}+B^{(\mathrm{c})}) + K_1W^{(\mathrm{bn})}B^{(\mathrm{bn})} \nonumber \\
%  &=& F^{(\mathrm{in})} \bullet K_1W^{(\mathrm{bn})}W^{(\mathrm{c})} + (K_1W^{(\mathrm{bn})}B^{(\mathrm{c})}+ K_1W^{(\mathrm{bn})}B^{(\mathrm{bn})}) \nonumber \\
%  &=& F^{(\mathrm{in})} \bullet W^{(\mathrm{c+bn})}+B^{(\mathrm{c+bn})}
%  \label{eq-ibn2}
%\end{eqnarray}
%\setlength{\arraycolsep}{5pt}

\setlength{\arraycolsep}{0.0em}
\begin{eqnarray}
    F^{(\mathrm{cat})} &=& \mathbf{CAT}(F^{(\mathrm{conv})},F^{(\mathrm{conv}+\mathrm{bn})})  \nonumber \\
    &=& F^{(\mathrm{in})} \bullet \mathbf{CAT}(W^{(\mathrm{c+bn})}) + \mathbf{CAT}(B^{(\mathrm{c+bn})}) \nonumber \\
    &=& F^{(\mathrm{in})} \bullet W^{(\mathrm{cat})} + B^{(\mathrm{cat})}
  \label{eq-convcat}
\end{eqnarray}
\setlength{\arraycolsep}{5pt}

The numbers of $F^{(\mathrm{conv})}$ and $F^{(\mathrm{conv}+\mathrm{bn})}$ to be concatenated may vary according to the specific implementation and could be 0 if necessary.

\setlength{\arraycolsep}{0.0em}
\begin{eqnarray}
    F^{(\mathrm{rep})} &=& \mathbf{CONV}_{1\times 1}(F^{(\mathrm{cat})}) \nonumber \\
    &=& (F^{(\mathrm{in})} \bullet W^{(\mathrm{cat})}+B^{(\mathrm{cat})}) \bullet W^{(\mathrm{c})}+B^{(\mathrm{c})} \nonumber \\
    &=& F^{(\mathrm{in})} \bullet \mathbf{MATMUL}(W^{(\mathrm{cat})},W^{(\mathrm{c})}) + \mathbf{MATMUL}(B^{(\mathrm{cat})}, W^{(\mathrm{c})}) + B^{(\mathrm{c})} \nonumber \\
    &=& F^{(\mathrm{in})} \bullet W^{(\mathrm{rep})} + B^{(\mathrm{rep})}
  \label{eq-convrep}
\end{eqnarray}
\setlength{\arraycolsep}{5pt}

$\mathbf{MATMUL}$ means matrix multiplication. $W^{(\mathrm{c})}$ and $W^{(\mathrm{c+bn})}$ both have the shape of $C^{(\mathrm{out})}\times C^{(\mathrm{in})} \times K_H \times K_W$, while $W^{(\mathrm{c+cat})}$ after concatenation has the shape of $NRC^{(\mathrm{out})} \times C^{(\mathrm{IN})} \times K_H \times K_W$ where $N$ is the number of branches and $K_H$ and $K_W$ are the kernel size. As the input feature and output feature of this ConvRep block are designed to have $C^{(\mathrm{in})}$ and $C^{(\mathrm{out})}$ channels, the final $\mathrm{CONV}_{1\times 1}$ after concatenation should have weight matrix with the shape of $C^{(\mathrm{in})}\times NRC^{(\mathrm{out})} \times 1 \times 1$. By squeezing, permuting, and matrix multiplication operations, the reparameterized weight matrix $W^{(\mathrm{rep})}$ and bias matrix $B^{(\mathrm{rep})}$ could be obtained as shown by Eq. \ref{eq-convrep}. 

\section{Two-branch feature fusion}
\label{sec:app-fusion}

After the re-parameterization, the two-branch structure could be converted back to two parallel convolution blocks and later on be fused by a multiplication operation as activation. The output fusion by $\mathbf{QCU}$ is represented by Eq. \ref{eq-branch-out}, and each branch can be represented as equations below Eq.\ref{eq-branch-rep-hat} and Eq. \ref{eq-branch-rep-til} as $\hat{I}$ and $\tilde{I}$. $B\in \mathbb{R}^{N \times C \times 1 \times 1}$ is the learnable bias in $\mathbf{QCU}$.

\begin{equation}
  I = \hat{I} \times \tilde{I} + B
  \label{eq-branch-out}
\end{equation}

For the convolution with kernel size of $(k_m, k_n)$, $k^{(h)}_m$ and $k^{(h)}_n$ are defined as $\frac{1}{2}(k_m-1)$ and $\frac{1}{2}(k_n-1)$. $l$ is the index of channels for input and output, while $L$ represents the total number of channels.

\begin{equation}
  \hat{I}^{l_{out}}_{i,j} = \hat{b}^{l_{out}}+\sum^{L_{in}}_{\hat{l}_{in}=0} \sum^{k^{(h)}_m}_{\hat{m}=-k^{(h)}_m} \sum^{k^{(h)}_n}_{\hat{n}=-k^{(h)}_n} \big( \hat{I}^{l_{in}}_{i+\hat{m},j+\hat{n}}\times \hat{K}^{l_{out},l_{in}}_{\hat{m},\hat{n}} \big)
  \label{eq-branch-rep-hat}
\end{equation} 

\begin{equation}
  \tilde{I}^{l_{out}}_{i,j} = \tilde{b}^{l_{out}}+\sum^{L_{in}}_{\tilde{l}_{in}=0} \sum^{k^{(h)}_m}_{\tilde{m}=-k^{(h)}_m} \sum^{k^{(h)}_n}_{\tilde{n}=-k^{(h)}_n} \big( \tilde{I}^{l_{in}}_{i+\tilde{m},j+\tilde{n}}\times \tilde{K}^{l_{out},l_{in}}_{\tilde{m},\tilde{n}} \big) 
  \label{eq-branch-rep-til}
\end{equation}

So that, the output feature value for a certain output channel in a fixed position $(i,j)$ can be represented precisely as Eq. \ref{eq-branch-mul} which could be further expressed as Eq. \ref{eq-branch-mul-exp}.

\begin{equation}
  I^{l_{out}}_{i,j} = \hat{I}^{l_{out}}_{i,j} \times \tilde{I}^{l_{out}}_{i,j} + b_{i,j}
  \label{eq-branch-mul}
\end{equation}

\setlength{\arraycolsep}{0.0em}
\begin{eqnarray}
I^{l_{out}}_{i,j} &=& b_{i,j} + \hat{b}^{l_{out}}\tilde{b}^{l_{out}} + \sum^{L_{in}}_{l_{in}=0} \sum^{k^{(h)}_m}_{m=-k^{(h)}_m} \sum^{k^{(h)}_n}_{n=-k^{(h)}_n} \big( \hat{I}^{l_{out}}_{i,j}\hat{K}^{l_{out},l_{in}}_{\hat{m},\hat{n}}\tilde{b}^{l_{out}} + \tilde{I}^{l_{out}}_{i,j}\tilde{K}^{l_{out},l_{in}}_{\tilde{m},\tilde{n}}\hat{b}^{l_{out}} \big) + \nonumber\\
  & & \sum^{L_{in}}_{\tilde{l}_{in}=0} \sum^{L_{in}}_{\hat{l}_{in}=0} \sum^{k^{(h)}_m}_{\tilde{m}=-k^{(h)}_m} \sum^{k^{(h)}_m}_{\hat{m}=-k^{(h)}_m} \sum^{k^{(h)}_n}_{\tilde{n}=-k^{(h)}_n} \sum^{k^{(h)}_n}_{\hat{n}=-k^{(h)}_n} \big( I^{\tilde{l}_{in}}_{i+\tilde{m},j+\tilde{n}} \times K^{l_{out}\tilde{l}_{in}}_{\tilde{m},\tilde{n}} \times I^{\hat{l}_{in}}_{i+\hat{m},j+\hat{n}} \times K^{l_{out}\hat{l}_{in}}_{\hat{m},\hat{n}} \big)
  \label{eq-branch-mul-exp}
\end{eqnarray}
\setlength{\arraycolsep}{5pt}

Consequently, the general formulation of the output feature expression obtained by the re-parameterized two-branch structure shall be $B + (\hat{K}+\tilde{K})I + KI^2$ which is in quadratic form rather than the linear form of $B + KI$. So, the capability of feature representation by this structure could be enhanced. 

However, the expression that could be converted into the form of $(\hat{K}\hat{I}+\hat{B})\times (\tilde{K}\tilde{I}+\tilde{B})$ have a strict constraint in feature representation pattern, which means the representation must have two sets of fixed values in solution space that are $(-\hat{I}/\hat{B},0)$ and $(-\tilde{I}/\tilde{B},0)$. Hence in order to solve this issue, an additional learnable bias $B$ is merged into the network to boost the ability to learn features.

\section{Theoretical analysis about different fusion methods on a toy example}
\label{sec:app-compfusion}

As mentioned in the paper, we compared different fusion methods including ADD(element-wise addition), CAT+CONV(concatenation plus convolution), MUL(element-wise multiplication), and the proposed QCU(Quadratic Connection Unit). As presented by Fig. \ref{fusion-comp} and Table \ref{table-abs}, QCU gains higher PSNR than these techniques. It could be roughly explained by that the mathematical expression of QCU shows more powerful representative capability than other fusion methods.

The analysis is conducted on a toy example, which is a simple two-branch network with input tensor $X$ of shape $\mathrm{N}\times\mathrm{C}\times\mathrm{H}\times\mathrm{W}$, output tensor $Y$, and two convolutions in each branch $\mathrm{CONV}^{\mathrm{(c)}}$ and $\mathrm{CONV}^{\mathrm{(s)}}$. Assuming the input channels and output channels are all equal to $\mathrm{C}$, and consequently, we have the weights and biases of the two convolutions, which are $W^{(\mathrm{c})}$, $B^{(\mathrm{c})}$, $W^{(\mathrm{s})}$ and $B^{(\mathrm{s})}$, all with $\mathrm{C}$ channels. For CAT+CONV method, we have one extra convolution $\mathrm{CONV}^{(\mathrm{cat})}$ with input channel $\mathrm{2C}$, output channel $\mathrm{C}$, weight $W^{(\mathrm{cat})}=[W^{(\mathrm{catc})}, W^{(\mathrm{cats})}]$ and bias $B^{(\mathrm{cat})}=[B^{(\mathrm{catc})}, B^{(\mathrm{cats})}]$. $W^{(\mathrm{cat})}$ has shape $\in\mathrm{C}\times\mathrm{2C}\times\mathrm{K}\times\mathrm{K}$, while $W^{(\mathrm{catc})}$ and $W^{(\mathrm{cats})}$ have shape $\in\mathrm{C}\times\mathrm{C}\times\mathrm{K}\times\mathrm{K}$, in which $\mathrm{K}$ is the kernel size.

The model could be further simplified by setting $\mathrm{C}=1$, and kernel size to be 1 as well. The channel and kernel assumption could convert the convolution into simple linear transformation.

\setlength{\arraycolsep}{0.0em}
\begin{eqnarray}
    Y^{\mathrm{(ADD)}} &=& (W^{(\mathrm{c})}X + B^{(\mathrm{c})}) + (W^{(\mathrm{s})}X + B^{(\mathrm{s})})  \nonumber \\
    &=& (W^{(\mathrm{c})}+W^{(\mathrm{s})})X + (B^{(\mathrm{c})} + B^{(\mathrm{s})})
  \label{eq-fusion-add}
\end{eqnarray}
\setlength{\arraycolsep}{5pt}

\setlength{\arraycolsep}{0.0em}
\begin{eqnarray}
    Y^{\mathrm{(CAT)}} &=& \mathbf{CAT}(W^{\mathrm{(c)}}X + B^\mathrm{(c)}, W^{\mathrm{(s)}}X + B^\mathrm{(s)})
  \label{eq-fusion-cat}
\end{eqnarray}
\setlength{\arraycolsep}{5pt}

\setlength{\arraycolsep}{0.0em}
\begin{eqnarray}
    Y^{\mathrm{(CAT+CONV)}} &=& W^{\mathrm{(cat)}}Y^{\mathrm{(CAT)}} + B^{\mathrm{(cat)}} \nonumber \\
    &=& W^{\mathrm{(catc)}}(W^{\mathrm{(c)}}X + B^{\mathrm{(c)}}) + B^{\mathrm{(catc)}} + W^{\mathrm{(cats)}}(W^{\mathrm{(s)}}X + B^{\mathrm{(s)}}) + B^{\mathrm{(cats)}} \nonumber \\
    &=& (W^{\mathrm{(catc)}}W^{\mathrm{(c)}} + W^{\mathrm{(cats)}}W^{\mathrm{(s)}})X + (W^{\mathrm{(catc)}}B^{\mathrm{(c)}} + B^{\mathrm{(catc)}} + W^{\mathrm{(cats)}}B^{\mathrm{(s)}} + B^{\mathrm{(cats)}})
  \label{eq-fusion-catconv}
\end{eqnarray}
\setlength{\arraycolsep}{5pt}

\setlength{\arraycolsep}{0.0em}
\begin{eqnarray}
    Y^{\mathrm{(MUL)}} &=& (W^{(\mathrm{c})}X + B^{(\mathrm{c})}) \times (W^{(\mathrm{s})}X + B^{(\mathrm{s})}) \nonumber \\
    &=& W^{\mathrm{(c)}}W^{\mathrm{(s)}}X^2 + (W^{\mathrm{(c)}}B^{\mathrm{(s)}} + W^{\mathrm{(s)}}B^{\mathrm{(c)}})X + B^{\mathrm{(c)}}B^{\mathrm{(s)}}
  \label{eq-fusion-mul}
\end{eqnarray}
\setlength{\arraycolsep}{5pt}

\setlength{\arraycolsep}{0.0em}
\begin{eqnarray}
    Y^{\mathrm{(QCU)}} &=& (W^{(\mathrm{c})}X + B^{(\mathrm{c})}) \times (W^{(\mathrm{s})}X + B^{(\mathrm{s})}) + B^{(\mathrm{QCU})} \nonumber \\
    &=& W^{\mathrm{(c)}}W^{\mathrm{(s)}}X^2 + (W^{\mathrm{(c)}}B^{\mathrm{(s)}} + W^{\mathrm{(s)}}B^{\mathrm{(c)}})X + (B^{\mathrm{(c)}}B^{\mathrm{(s)}} + B^{(\mathrm{QCU})})
  \label{eq-fusion-qcu}
\end{eqnarray}
\setlength{\arraycolsep}{5pt}

Comparing the expressions of all the fusion methods, the representative capability of these methods could be summarized. As indicated by Table \ref{table-abs}, the performance on PSNR by the fusion methods is ranked as QCU $>$ MUL $>$ CAT+CONV $>$ ADD. It is observed that QCU and MUL both have an extra second-order term $X^2$, which makes QCU and MUL have the more powerful representative capability to fit more complex models. On the other hand, CAT+CONV could be regarded as a weighted addition operation, in which the weights of $W^{\mathrm{(c)}}$, $W^{\mathrm{(s)}}$, $B^{\mathrm{(c)}}$, and $B^{\mathrm{(s)}}$ are all adjusted by the following convolution $\mathbf{CONV}^{\mathrm{(cat)}}$. Thus, CAT+CONV fusion could generate more complex fitting. It should be noted that those tricks could significantly improve the performance of small networks. But in large models, since the networks already have very large depth and width to fit complex distributions, those tricks might not be necessarily helpful.

\section{Comparisons with models in SR task with scale factor $\times 3$}
\label{sec:app-table-more-comp}

More comparisons for the scale factor of $\times 3$ are shown in Table \ref{table-more-sr}. We trained these models on DIV2K and tested them on Set5, Set14, BSD100, and Urban100 following Table \ref{table-sr}.

\begin{table*}[htbp]
\centering
\resizebox{1.0\columnwidth}{!}{
    \begin{tabular}{lcccccccccc}
    \toprule
                Method & Scale & \#P & Avg latency(ms) & FPS(2K) & Set5 & Set14 & BSD100 & BSD100 Score & Urban100 & Urban100 Score\\
    \midrule
                ABPN\cite{du2021anchorbased} & $\times 3$ & 53K & 59.3 & 17 & 32.99 & 29.46 & 28.48 & 2.264 & 26.34 & 2.434 \\
                SCSRN\cite{ignatov2022efficient} & $\times 3$ & 67K & 43.1 & 23 & \textbf{33.20} & \textbf{29.57} & \textbf{28.56} & 3.634 & \textbf{26.52} & 4.294 \\
                HOPN\cite{ignatov2022efficient} & $\times 3$ & 48K & 32.1 & 31 & 32.66 & 29.28 & 28.35 & 3.647 & 26.33 & 2.966 \\
    \hline
                SYENet (\textbf{Ours}) & $\times 3$ & 7.9K & \textbf{11.3} & 88 & 32.93 & 29.44 & 28.47 & \textbf{12.234} & 26.33 & \textbf{12.595} \\
    \hline
    \end{tabular}
}
\quad \\
    \caption{Comparison on super-resolution issue between the results by PSNR(dB), SSIM, and comprehensive score with SOTA: the models are ranked by score defined by Eq. \ref{eq-score} on BSD100 dataset. The normalization factors $C$ for BSD100 Score are $1.8\mathrm{E}16(\times 2)$, $1\mathrm{E}14(\times 3)$, and $5\mathrm{E}14(\times 4)$. While the normalization factors $C$ for Urban100 Score are $2.5\mathrm{E}15(\times 2)$, $3.5\mathrm{E}12(\times 3)$, and $1.6\mathrm{E}13(\times 4)$.}
    \label{table-more-sr}
\end{table*}

\section{Constraints of low-level vision tasks in mobile devices}
\label{sec:app-constraints}

\subsection{Constraints of input size and output size}

Low-level vision tasks are basically regression questions. The output of the model is at least the same size as the input, and particularly for super-resolution problems, the output size shall be many times larger than that of the input. However, the output of high-level vision tasks like classification questions could be a vector containing scores of different object labels, which is much smaller than the low-level vision tasks output. Apart from output size, the input to high-level vision tasks could also be significantly smaller than low-level vision task input. Normally, the input image would be resized into as small as $128\times 128$ or $256\times 256$. However, for low-level vision tasks, the information of pixels is rather vital and should not be down-scaled or degraded before processing.

\begin{table}[htbp]
%\vspace{-1mm}
\centering
\resizebox{1.0\columnwidth}{!}{
    \begin{tabular}{c|ccccccc}
    \hline
        Methods & Task & $\#P$ & Input & Output & MACs(G) & FLOPs(G) & Latency(Platform)\\
    \hline
        EfficientFormer\cite{li2022rethinking} & Classification & 37.1M & $224\times 224$ & - & 3.57 & - & 4.2ms(iPhone 12 NPU) \\
        LightViT-B\cite{huang2022lightvit} & Classification & 35.2M & $224\times 224$ & - & - & 3.9 & 1.2ms(V100 GPU) \\
        SYENet(Ours) & SR$\times 2$ & 4.9K & $960 \times 540$ & $1920 \times 1080$ & - & 2.83 & 16.5ms(Qualcomm Snapdragon mobile SoC) \\
    \hline
    \end{tabular}
}
\quad \\
%\vspace{-3mm}
    \caption{It is obviously shown from the comparison between SYENet and efficient networks for high-level vision task that low-level vision task is more challenging as it requires much fewer parameters to achieve real-time inference due to the large input and output size.}
    \label{table-comp-efsye}
%\vspace{-3mm}
\end{table}

Two typical examples of efficient high-level vision networks are EfficientFormer\cite{li2022rethinking} and LightViT\cite{huang2022lightvit}. With 37.1 million parameters, EfficientFormer Supernet could still achieve an extremely low latency of 4.2ms on iPhone 12 NPU. While, as shown in Table \ref{table-comp-efsye}, our proposed method with only 4.9k parameters could only reach a latency of 16.5ms on Qualcomm Snapdragon mobile SoC due to the large input and output size.

\subsection{Constraints of small memory bandwidth of mobile devices}

Information Multi-distillation Block IMDB\cite{Hui-IMDN-2019} is widely utilised in the efficient SR challenges\cite{zhang2020aim} by many participators as well as the winning teams in some years. However, the structure of IMDB involves a serious disadvantage for mobile devices as it requires the network to store the results generated by each step which should be concatenated after all the steps are done. Therefore, during the inference, once the memory is too small, and consequently data transfer between the memory and disk occurs, the latency shall be seriously enlarged. Thus, keeping memory usage small is another challenge for low-level vision tasks. This is also the reason that the structure of IMDB is not employed in SYENet.

\subsection{The influence of network depth towards network latency}

The tradeoff between the network depth and kernel size is shown in Table \ref{apdx:table-dk-tradeoff}. Apparently, one ${5\times 5}$ convolution requires more numerical calculations than two ${3\times 3}$ convolutions. However, as indicated by Table \ref{apdx:table-dk-tradeoff}, deeper networks have more data transfer operations between DRAM and local SRAM when the feature map size is too huge. Thus, the larger depth will also enlarge the latency as well. Sometimes, the influence of depth could be even larger than FLOPs in certain platforms due to the inconsistency between hardware acceleration policies and network structures. Therefore, taking the hardware acceleration policies into consideration is also vital for designing the network structure. 

\begin{table}[htbp]
\centering
\resizebox{0.6\columnwidth}{!}{
    \begin{tabular}{c|ccc}
    \hline
        Methods & PSNR & Latency(ms) & MAIISP Score \\
    \hline
        $\mathrm{CONV}_{5\times 5}$ & 24.8732 & \textbf{11.4} & \textbf{16.57} \\
        $\mathrm{CONV}_{3\times 3}\rightarrow \mathrm{CONV}_{3\times 3}$ & 24.8824 & 12.9 & 14.83 \\
        $\mathrm{CONV}_{3\times 3}\rightarrow \mathrm{PReLU}\rightarrow \mathrm{CONV}_{3\times 3}$ & \textbf{24.9349} & 13.4 & 15.35 \\
    \hline
    \end{tabular}
}
\quad \\
    \caption{Depth-kernel size tradeoff: tested on MAI ISP dataset}
    \label{apdx:table-dk-tradeoff}
\end{table}

\section{Disadvantage of SYENet on VSR task}

The experiment results of SYENet on Video SR problems are very unsatisfactory, which is even worse than that of SISR. As suggested by BasicVSR\cite{chan2021basicvsr}, the VSR network should have four modules which are propagation, alignment, aggregation, and upsampling. Since SYENet is too small to implement the propagation and alignment module, SYENet could not utilise the information of neighboring frames to enhance the VSR performance. In fact, the neighboring frames injected into SYENet actually impair the restoration as they provide misaligned pixel information that SYENet cannot make good use of. 

\section{More low-light enhancement comparisons}
\label{apdx:lol-comps}

\begin{figure*}[t]
  \centering
    \includegraphics[width=1.0\linewidth]{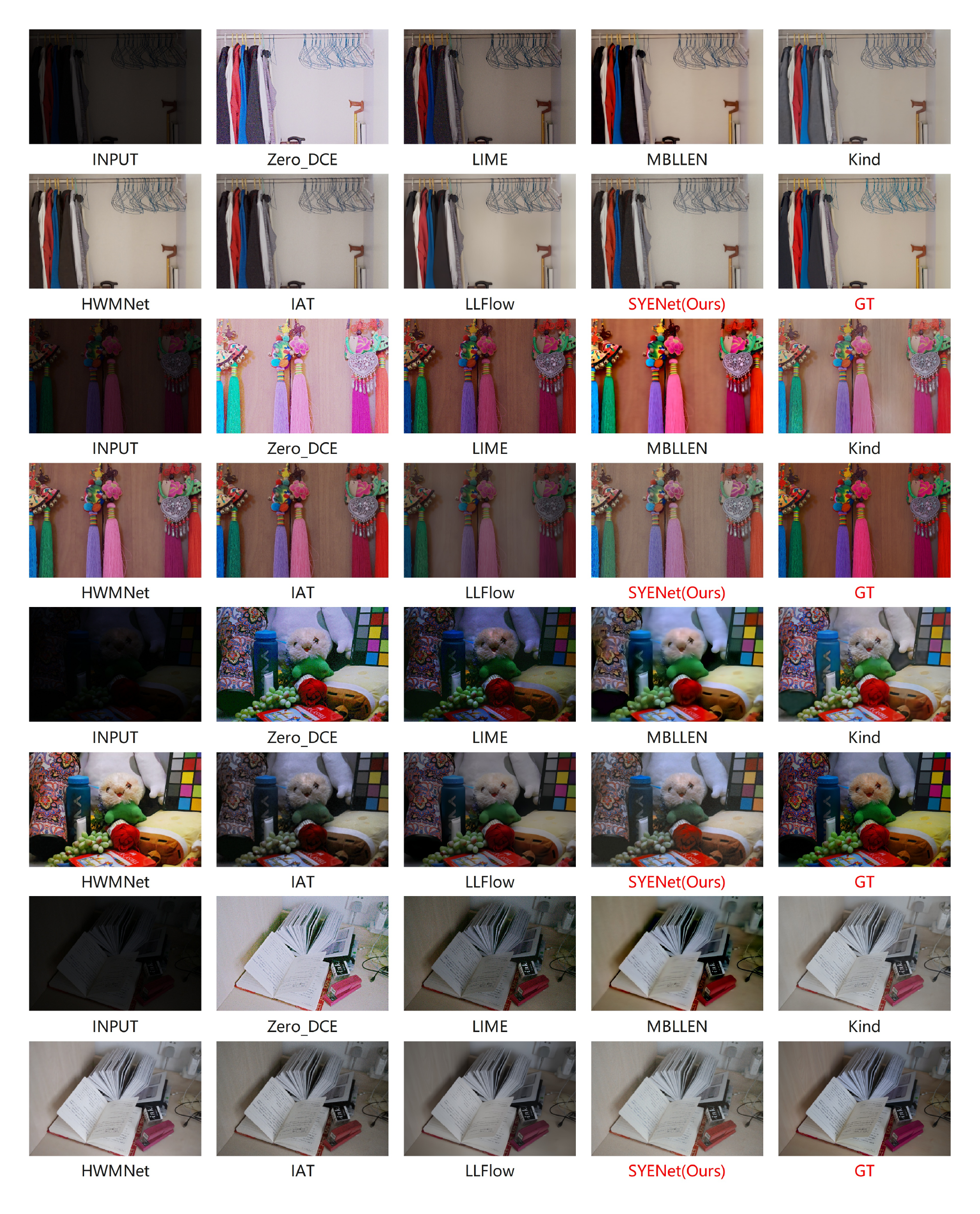}
   \caption{Low-light enhancement comparisons}
   \label{app-lle}
\end{figure*}

The images are shown in Fig. \ref{app-lle}.

\section{More image signal processing comparisons}
\label{apdx:isp-comp}

\begin{figure*}[t]
  \centering
    \includegraphics[width=1.0\linewidth]{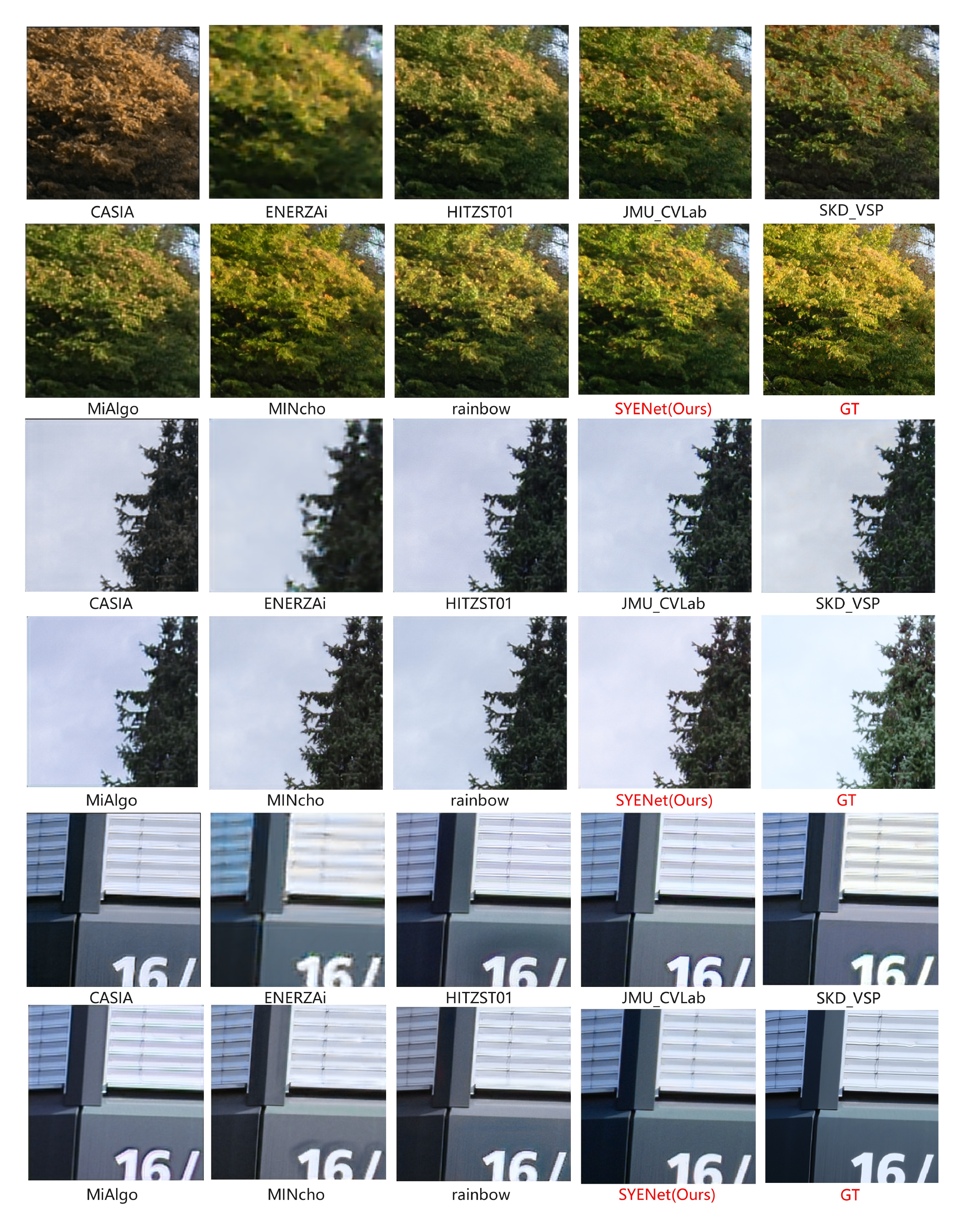}
   \caption{Image signal processing comparisons}
   \label{app-isp}
\end{figure*}

The images are shown in Fig. \ref{app-isp}.

\section{More Feature maps by each branch}
\label{apdx:fmap}

\begin{figure*}[t]
  \centering
    \includegraphics[width=0.95\linewidth]{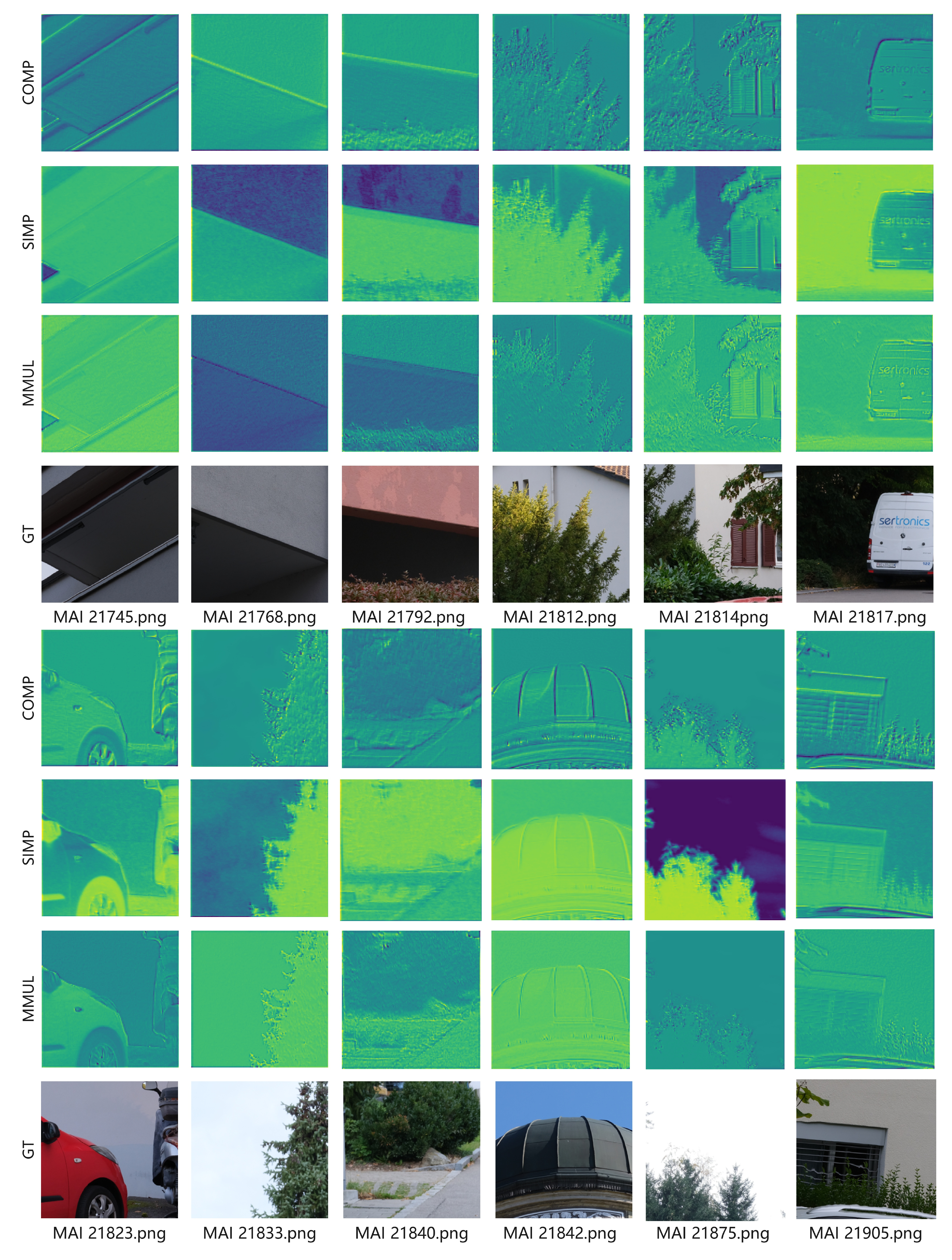}
   \caption{Feature maps of each branch and fused result}
   \label{app-fmap}
\end{figure*}

The images are shown in Fig. \ref{app-fmap}.

\clearpage

%{\small
%\bibliographystyle{ieee_fullname}
%\bibliography{egbib}
%}

\end{document}